\documentclass[11pt]{article}

\usepackage[preprint]{acl}

\usepackage{times}
\usepackage{latexsym}

\usepackage[T1]{fontenc}

\usepackage[utf8]{inputenc}

\usepackage{microtype}

\usepackage{inconsolata}

\usepackage{hyperref}
\usepackage{url}

\usepackage{graphicx}
\usepackage{subfigure}
\usepackage{wrapfig}

\usepackage{amsmath}
\usepackage{amstext}
\usepackage{amsfonts}
\usepackage{bm}

\usepackage{bbm}
\usepackage{multirow}
\usepackage{booktabs}
\usepackage{array}
\usepackage{caption}
\usepackage{multirow}
\usepackage{booktabs}
\usepackage{array}
\usepackage{caption}
\usepackage{color}
\usepackage{colortbl}
\usepackage[dvipsnames]{xcolor}
\usepackage{tablefootnote}
\usepackage{adjustbox}

\definecolor{mygray}{gray}{.9}
\definecolor{c0}{cmyk}{1,0.3968,0,0.2588} 

\usepackage{caption}
\usepackage{algorithm}
\usepackage{algpseudocode}

\newcommand{\mydarkcolor}[1]{\textcolor[RGB]{64,101,149}{#1}}
\newcommand{\gray}[1]{\textcolor{gray}{#1}}
\algnewcommand{\LineComment}[1]{\Statex ~~~~~~\textsc{//}~\textit{#1}}
\algnewcommand{\DoubleLineComment}[1]{\Statex ~~~~~~~~~~~~\textsc{//}~\textit{#1}}

\usepackage{enumitem}
\setenumerate[1]{itemsep=0pt,partopsep=0pt,parsep=\parskip,topsep=5pt}
\setitemize[1]{itemsep=0pt,partopsep=0pt,parsep=\parskip,topsep=5pt}
\setdescription{itemsep=0pt,partopsep=0pt,parsep=\parskip,topsep=5pt}
\definecolor{myblue}{HTML}{4285f4}

\usepackage{tcolorbox}
\tcbuselibrary{skins}
\definecolor{lightgreen}{RGB}{225, 239, 217} 
\usepackage{etoolbox}
\AtBeginEnvironment{tcolorbox}{\small}

\usepackage[mathscr]{euscript}

\usepackage{tikz}
\definecolor{myred}{RGB}{240, 119, 115} 
\definecolor{myblue}{RGB}{78, 149, 217} 
\definecolor{mygreen}{RGB}{78, 167, 46} 
\definecolor{delimiter}{RGB}{242, 242, 242} 
\newcommand{\blue}{\cellcolor{c0!5}} 
\newcommand{\green}{\cellcolor{mygreen!15}} 
\newcommand{\mybox}[2]{\tikz[baseline=(MeNode.base)]{\node[rounded corners=2pt, inner sep=2pt, fill=#1](MeNode){#2};}}

\usepackage{amssymb}  
\usepackage{pifont}

\usepackage{xspace}
\usepackage{fontawesome}
\newcommand{\method}{\textsc{ToolSpec}\xspace}

\usepackage{twemojis}

\usepackage{ulem}

\title{
\texorpdfstring{\includegraphics[width=18pt]{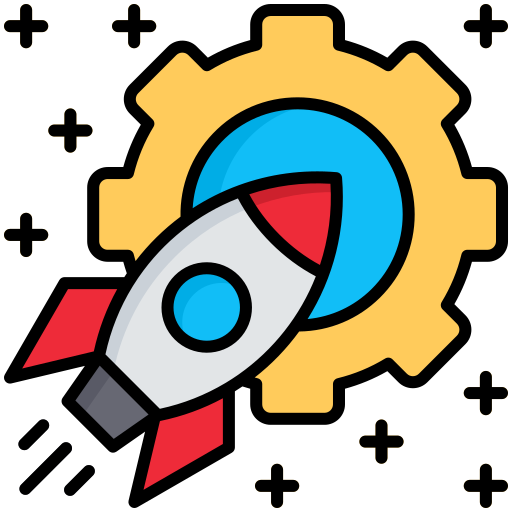}}{}
\method: Accelerating Tool Calling via Schema-Aware and Retrieval-Augmented Speculative Decoding}
\author{{Heming Xia}\textsuperscript{\rm 1}, {Yongqi Li}\textsuperscript{\rm 1}\thanks{\ Corresponding author.}, {Cunxiao Du}\textsuperscript{\rm 1}, {Mingbo Song}\textsuperscript{\rm 2}, {Wenjie Li}\textsuperscript{\rm 1}\\
  \textsuperscript{\rm 1} Department of Computing, The Hong Kong Polytechnic University 
  \textsuperscript{\rm 2} Peking University \\
  {\tt \{he-ming.xia\}@connect.polyu.hk; \{harukioj\}@outlook.com}
}

\begin{document}
\maketitle
\begin{abstract}
Tool calling has greatly expanded the practical utility of large language models (LLMs) by enabling them to interact with external applications. As LLM capabilities advance, effective tool use increasingly involves multi-step, multi-turn interactions to solve complex tasks. However, the resulting growth in tool interactions incurs substantial latency, posing a key challenge for real-time LLM serving. Through empirical analysis, we find that tool-calling traces are highly structured, conform to constrained schemas, and often exhibit recurring invocation patterns. Motivated by these observations, we propose \method, a schema-aware, retrieval-augmented speculative decoding method for accelerating tool calling. \method exploits predefined tool schemas to generate accurate drafts, using a finite-state machine to alternate between deterministic schema token filling and speculative generation for variable fields. In addition, \method retrieves similar historical tool invocations and reuses them as drafts to further improve efficiency. \method presents a training-free, plug-and-play solution that can be seamlessly integrated into existing LLM workflows. Experimental results across multiple tool-calling benchmarks demonstrate that \method achieves up to a $4.2\times$ speedup, substantially outperforming existing training-free speculative decoding methods. We release our code at \url{https://github.com/hemingkx/ToolSpec}.
\end{abstract}

\section{Introduction}

Recent advances in large language models (LLMs) have substantially broadened their capabilities, including open-domain dialogue, mathematical reasoning, and program synthesis~\citep{o1, sonnet3_5, Qwen3}. In parallel, tool calling has emerged as a pivotal capability that enables LLMs to interact with external tools such as search engines~\citep{Search-R1, deepresearcher}, calculators~\citep{Schick:2023toolformer, Qin:2024toolllm}, and code interpreters~\citep{PAL, Gou:2024Tora}. By incorporating external tools, LLMs can alleviate fundamental limitations of standalone models, including outdated knowledge and arithmetic errors~\cite{Qin2024:ToolSurvey}. As LLM capabilities continue to improve, tool use increasingly adopts multi-step, multi-turn, feedback-driven interaction loops to solve complex tasks~\cite{gorilla, yao2025taubench}.

\begin{figure}[t]
\centering
\includegraphics[width=0.95\columnwidth]{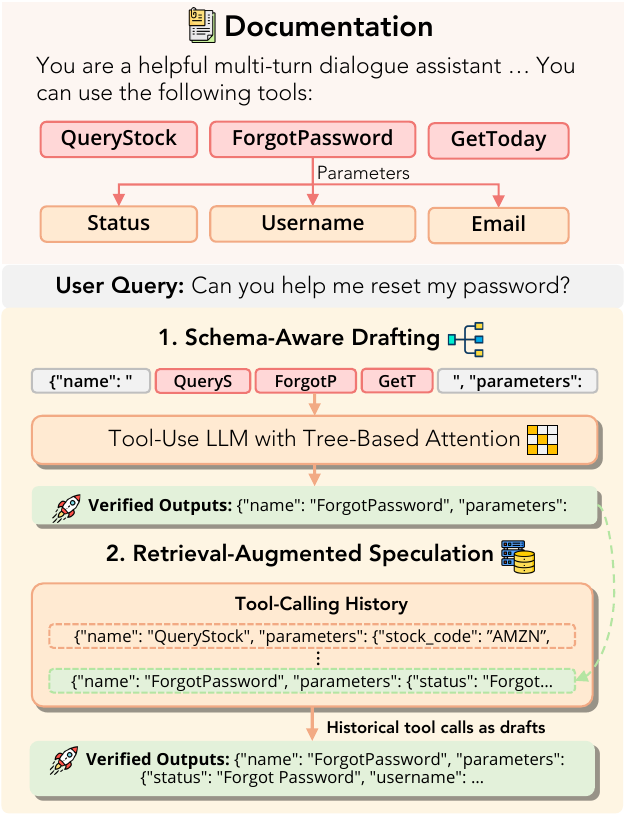}
\caption{Illustration of \method, which accelerates tool-calling generation via two innovations: \textbf{1) schema-aware drafting}, where predefined tool schemas serve as faithful drafts and enable parallel verification of constrained variables (e.g., \colorbox{pink}{tool names}); and \textbf{2) retrieval-augmented speculation}, retrieving and reusing similar historical tool invocations as high-quality drafts.}
\label{fig:intro}
\end{figure}

However, the growing frequency of tool invocations substantially increases inference latency, posing a major challenge for real-time LLM serving. Recent work has proposed several strategies to accelerate tool calling~\cite{LLMCompiler, zhu2025:dividethenaggregate, xu2024conveyor, nichols2025spec}. One line of research models tool-calling trajectories as directed acyclic graphs (DAGs) that capture inter-tool dependencies and then execute independent tools in parallel to reduce overall latency~\cite{LLMCompiler, zhu2025:dividethenaggregate}. Another line improves parallelism by overlapping tool execution with text generation, that is, launching tool calls while the model is still decoding, thereby further accelerating end-to-end tool-calling~\cite{xu2024conveyor, nichols2025spec}. 

Despite these advances in accelerating \textit{tool execution}, this work shifts focus to improving the efficiency of \textit{tool-calling generation}. Our preliminary analysis indicates that generation latency can be a dominant bottleneck: on the widely used benchmark ToolBench~\cite{Qin:2024toolllm}, generating tool calls with Qwen2.5-14B-Instruct accounts for approximately $80\%$ of the end-to-end inference latency and is roughly $4\times$ larger than the time spent executing the tools. Moreover, whereas tool execution time is often relatively stable for a fixed toolset and environment, tool-calling generation latency grows with model size and with the length of the generated tool-call sequence, making it the primary contributor to latency in many tool-using scenarios.

In this work, we present two key observations regarding tool calling: \textbf{1) tool-calling outputs are highly structured.} These outputs are typically expressed in JSON and adhere to strictly constrained schemas. As a result, among the generated tokens, only a small subset corresponds to variable fields, while the rest are schema-determined; \textbf{2) LLMs exhibit recurring tool-call patterns across different requests,} indicating the presence of reusable structural and semantic regularities. Motivated by these insights, we propose \method, a schema-aware and retrieval-augmented speculative decoding (SD) approach for accelerating tool calling. \method leverages the task-specific tool schemas to construct high-quality draft sequences, and employs a finite-state machine to alternate between deterministic schema token filling and speculative generation for variable fields. Furthermore, \method retrieves similar historical tool calls and reuses them as draft candidates, thereby further improving efficiency. Experiments on widely used tool-calling benchmarks, including API-Bank, ToolAlpaca, BFCLv2, and ToolBench, demonstrate that \method achieves up to a $4.2\times$ speedup and outperforms existing SD methods.

We summarize our contributions as follows:
\begin{itemize}
    \item We propose \method, a schema-aware and retrieval-augmented SD approach for accelerating tool-calling generation. It presents a \textit{training-free} solution that can be seamlessly integrated into existing LLM workflows.
    \item Extensive experiments across diverse models and tasks demonstrate that \method consistently achieves a $3.5\times$$\sim$$4.2\times$ speedup, with up to a $71\%$ relative improvement over the previous state-of-the-art SD method.
    \item Further analysis validates the effectiveness of our proposed components and highlights the impact of format adherence on \method’s speedup, indicating its efficiency potential as LLM capabilities continue to advance.
\end{itemize}
\section{Related Work}

\paragraph{Efficient Tool Calling}
As large language models (LLMs) increasingly rely on multi-step, multi-turn tool-calling workflows, inference latency in LLM serving has grown substantially. To address this, recent studies leverage directed acyclic graphs (DAGs) to model inter-tool dependencies, enabling parallel execution of independent API calls and thereby reducing latency~\cite{LLMCompiler, zhu2025:dividethenaggregate}. Another line of work executes tool calls proactively during generation to minimize system idle time~\citep{xu2024conveyor, nichols2025spec}. In contrast to prior methods that focus on tool execution, this work shifts attention to accelerating tool-calling generation, further improving the inference efficiency of tool-use LLMs.

\paragraph{Speculative Decoding}
In recent years, speculative decoding (SD) has emerged as a widely adopted paradigm for accelerating LLM inference~\cite{leviathan:2023spec, xia:2023specdec}. At each decoding step, SD first generates a sequence of draft tokens, which are subsequently verified in parallel by the target LLM~\citep{xia:2024survey}. Recent advances in SD, such as the Eagle series~\citep{eagle, li2024:eagle2, li2025:eagle3}, introduce lightweight draft models layered on top of the target LLM, achieving promising speedups. While these approaches require additional training or parameters, an alternative line of research focuses on training-free SD methods~\citep{luo:2025tokenrecycle, saxena2023pld, SWIFT}. Notably, Token Recycling~\citep{luo:2025tokenrecycle} stores the top-$k$ candidate tokens of previous steps in an adjacency matrix and uses breadth-first search to construct a draft tree. Similarly, SAM-Decoding~\citep{hu2025:samd} leverages a suffix automaton to efficiently retrieve high-quality drafts from both the input context and a static text corpus.
\section{Preliminary Study}
\label{sec:preliminary}

In this section, we conduct an empirical analysis of tool-calling generation, covering (i) the latency breakdown between generation and tool execution, and (ii) analysis of tool-call patterns.

\begin{figure}[t]
\centering
\vspace{-0.5cm}
\includegraphics[width=0.95\columnwidth]{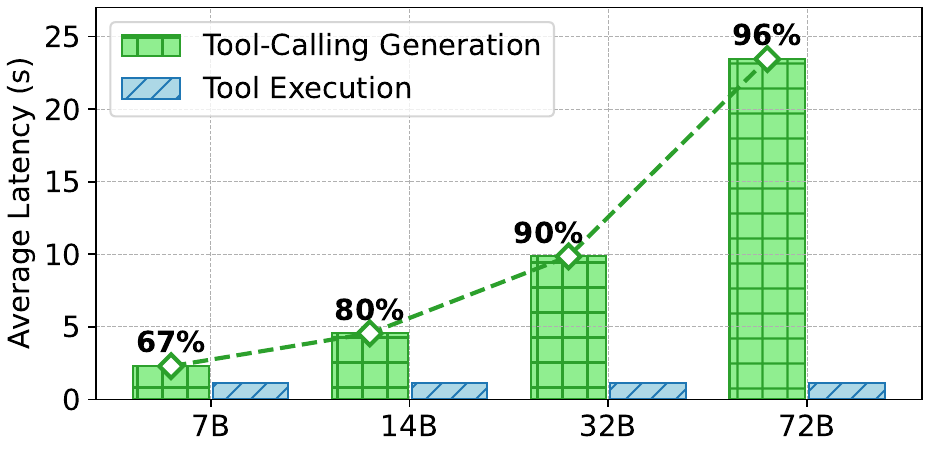}
\caption{Latency breakdown of the Qwen2.5-Instruct model series on ToolBench~\citep{Qin:2024toolllm}. Tool execution latency remains essentially constant, whereas tool-calling generation latency increases with model scale, reaching up to $\bf 96\%$ of the end-to-end latency.}
\label{fig:latency_breakdown}
\end{figure}

\subsection{Latency Breakdown}
We begin by analyzing the latency bottlenecks in tool calling. Experiments are conducted using the Qwen2.5-Instruct series on ToolBench~\citep{Qin:2024toolllm}. As shown in Figure~\ref{fig:latency_breakdown}, tool execution latency remains nearly constant across model scales, whereas tool-calling generation latency increases significantly, accounting for up to $\bf{96\%}$ of the end-to-end latency for Qwen2.5-72B-Instruct. These results indicate that tool-calling generation is the dominant bottleneck in tool-calling pipelines, particularly in multi-tool, multi-turn scenarios, underscoring the importance of accelerating generation.

\subsection{Analysis of Tool-Call Patterns}
\label{sec:pattern_analysis}

\begin{figure}[t]
\centering
\vspace{-0.5cm}
\includegraphics[width=0.95\columnwidth]{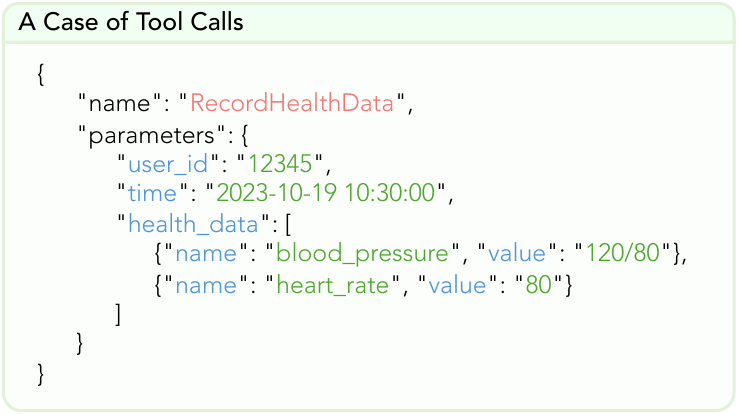}
\caption{A tool-calling trace from API-Bank~\citep{apibank}, with different fields highlighted in color.}
\label{fig:case}
\end{figure}

Figure~\ref{fig:case} shows an example tool-calling trace from API-Bank~\citep{apibank}. As illustrated, these traces are highly structured JSON objects that conform to a constrained schema, consisting of:

\begin{itemize}
    \item \textbf{schema tokens}, e.g., \texttt{``name''}, \texttt{``parameters''}, and punctuation, which form the fixed scaffolding required by the tool-call schema;
    \item \textbf{tool and parameter names}, highlighted in \textcolor{myred}{red} and \textcolor{myblue}{blue}, respectively, which specify the target tool and its arguments;
    \item \textbf{parameter values}, in \textcolor{mygreen}{green}, which instantiate the arguments with query-specific content.
\end{itemize}

These structured patterns distinguish tool-calling generation from general natural-language generation, suggesting that it can be formulated as a \textit{constrained decoding} problem. In practice, valid tool calls must adhere to strict JSON schemas, implying that a large portion of the output tokens, i.e., schema tokens, is effectively predetermined. Once the target tool is selected, its parameter names are drawn from a finite, schema-defined set (e.g., \texttt{``user\_id''}, \texttt{``time''}, and \texttt{``health\_data''}). Consequently, the model only needs to generate a small subset of tokens, i.e., the parameter values, conditioned on the user query and the dialogue context.

In the following, we further analyze practical tool-calling behaviors in LLMs, including schema adherence and recurring tool-invocation patterns.

\begin{table}[t]
\centering
\small
\begin{tabular}{@{}lcc@{}}
\toprule
\bf Models & \bf API-Bank & \bf ToolAlpaca \\\midrule
Llama-3.1-8B-Instruct &$100.0\%$ &$100.0\%$\\
Llama-3.2-3B-Instruct &$99.5\%$ &$93.4\%$\\
Qwen2.5-14B-Instruct &$99.7\%$ &$100.0\%$\\
Qwen2.5-7B-Instruct &$100.0\%$ &$100.0\%$\\
\bottomrule
\end{tabular}
\caption{Format adherence of different models.}
\label{tab:format_adherence}
\end{table}

\paragraph{Schema Adherence}
As shown in Table~\ref{tab:format_adherence}, current tool-use LLMs exhibit strong adherence to tool-calling formats, i.e., their outputs can be reliably parsed as valid JSON. Both the LLaMA-Instruct and Qwen2.5-Instruct model families achieve over $99\%$ accuracy in following the required tool-call schema. This observation further supports our earlier claim that tool-calling generation can be formulated as constrained decoding under predefined schemas. In light of this, acceleration methods designed for general text generation may be suboptimal for tool calling. In contrast, explicitly exploiting the constrained structure of tool-calling generation can provide additional efficiency gains.

\begin{table}[t]
\centering
\vspace{-0.5cm}
\small
\begin{tabular}{@{}lccc@{}}
\toprule
\bf Datasets & \bf \#Tools & \bf \#Instances & \bf Avg. Tool Rep. \\\midrule
API-Bank &2138  &5221 & 10.95 \\
ToolAlpaca &426 &3938 & 4.82 \\
BFCLv2 & 739 &2251  & 9.01 \\
\bottomrule
\end{tabular}
\caption{Statistics of tool-calling benchmarks.}
\label{tab:tool_frequency}
\end{table}

\paragraph{Repetitive Tool Calls}
Beyond schema adherence, we observe that LLMs frequently invoke the same tools to satisfy recurring user intents and preference patterns. This behavior is consistently observed across widely used tool-calling benchmarks. As shown in Table~\ref{tab:tool_frequency}, repeated invocations of identical tools occur frequently in API-Bank~\citep{apibank}, with the average number of calls per tool exceeding $10.95$. This observation motivates exploiting historical tool-calling traces. Rather than discarding prior calls, a system can reuse previously issued tool calls when they remain valid, thereby further improving tool-calling efficiency.
\section{\method}
\label{sec:toolspec}
In this section, we introduce \method, a plug-and-play SD approach for accelerating tool-calling generation. The method leverages the highly structured nature of tool calls and recurring historical patterns to construct high-quality drafts with negligible overhead. As shown in Figure~\ref{fig:toolspec}, \method integrates two complementary drafting strategies: \textit{schema-aware drafting} and \textit{retrieval-augmented speculation}, which we detail below.

\subsection{Task Formulation}
Given a user query and an input context containing system prompts and tool documentation, \method extracts tool specifications from the documentation and converts them into a structured schema. The schema defines (i) the set of schema tokens $\Sigma_{\text{schema}}$, (ii) a set of callable tools $\mathcal{C}_{\text{tools}}$ (e.g., \texttt{ForgotPassword}), and (iii) the parameter fields associated with each tool (e.g., \texttt{Status}, \texttt{Username}).

Let $\mathcal{M}$ denote the target model, $\boldsymbol{x}$ the input context, and $\boldsymbol{y} = (y_1, y_2, \ldots, y_L)$ the tool-call sequence to be generated. At decoding step $t$, following the standard SD paradigm, \method first proposes a sequence of draft tokens $\tilde{\boldsymbol{y}}_{t:t+\gamma-1} = (\tilde{y}_{t}, \tilde{y}_{t+1}, \ldots, \tilde{y}_{t+\gamma-1})$, as predictions of the subsequent decoding steps. The target model $\mathcal{M}$ then verifies these draft candidates in parallel and accepts the longest prefix that is consistent with its probability distribution. When the drafting process incurs significantly lower latency than verification, maximizing the acceptance length of $\tilde{\boldsymbol{y}}$ directly improves the effective decoding parallelism and reduces the end-to-end latency.

\begin{figure}[t]
\centering
\vspace{-0.5cm}
\includegraphics[width=0.95\columnwidth]{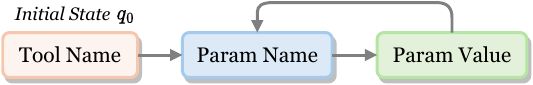}
\caption{Illustration of state transitions in \method. Once tool calling is triggered by \texttt{<tool\_call>}, the FSM enters the tool-name state $q_t$, and then alternates between the parameter-name state $q_p$ and parameter-value state $q_v$ until the tool call is completed.}
\label{fig:transition}
\end{figure}

\subsection{Schema-aware Drafting}
As discussed in Section~\ref{sec:pattern_analysis}, tool invocations from advanced LLMs adhere to predefined schema formats. This structural regularity enables deterministic drafting for large portions of the output. To exploit this property, \method constructs a finite-state machine (FSM) derived from the parsed tool schema. Formally, the FSM is defined as
\begin{equation}
\mathcal{F} = (\mathcal{Q}, \Sigma_{\text{schema}}, \delta, q_0),
\end{equation}
where $\mathcal{Q}$ is the set of states, $\Sigma_{\text{schema}}$ is the schema token alphabet, $\delta: \mathcal{Q} \times \Sigma_{\text{schema}} \rightarrow \mathcal{Q}$ is the transition function, and $q_0\in \mathcal{Q}$ is the initial state.

The state space $\mathcal{Q}$ decomposes the tool-calling generation process into structural components:
\begin{itemize}
\item $q_t$: \textbf{tool-name state}, where the model selects a tool from the predefined tool set $\mathcal{C}_{\text{tools}}$.
\item $q_p$: \textbf{parameter-name state}, where the model selects one parameter field of the chosen tool.
\item $q_v$: \textbf{parameter-value state}, where the model generates the value of the selected parameter.
\item $q_o$: \textbf{``others'' state}, corresponding to general natural language generation outside tool calls.
\end{itemize}

\begin{figure*}[t]
\centering
\vspace{-1.0cm}
\includegraphics[width=0.95\textwidth]{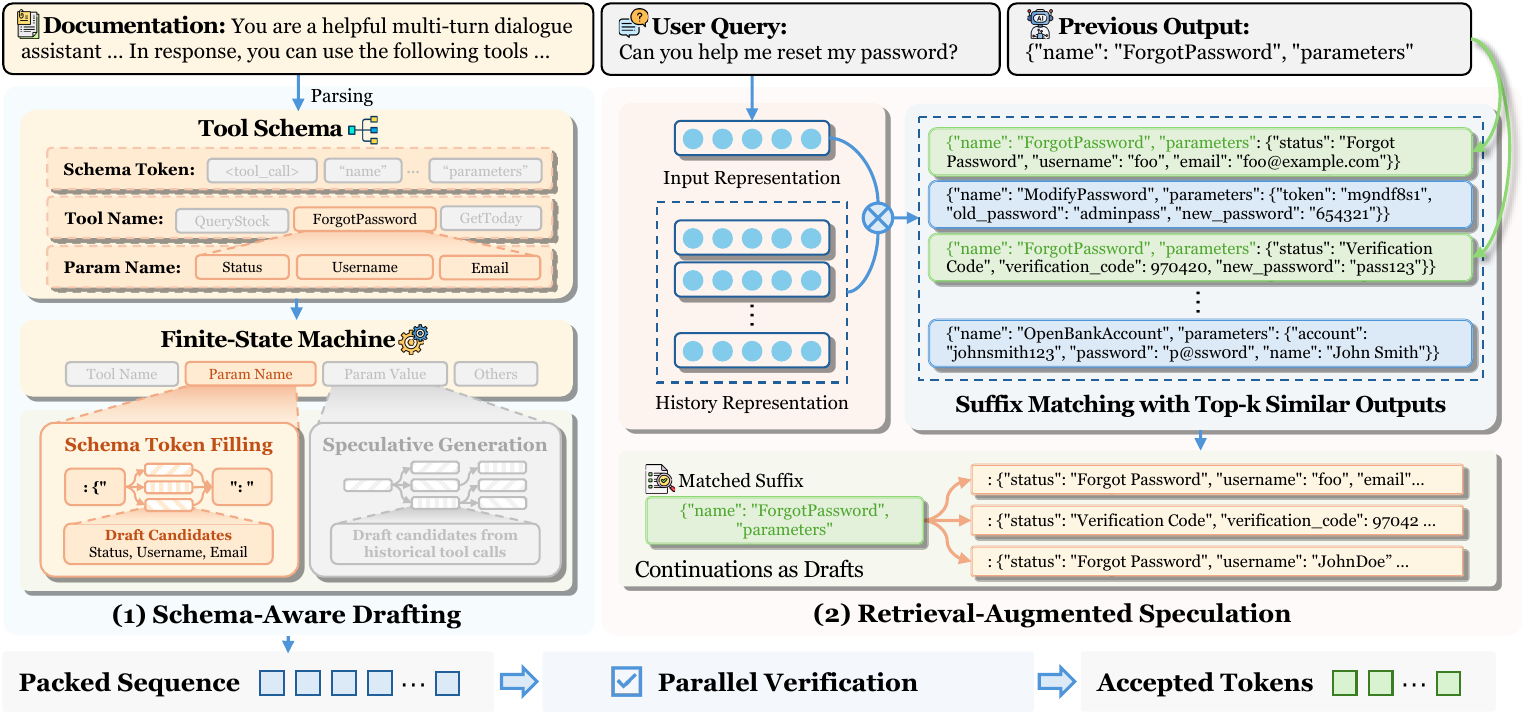}
\caption{Illustration of tool-calling generation with \method. To accelerate structural tool calling, \method integrates two complementary drafting strategies: \textbf{(a) schema-aware drafting}, where a finite-state machine guides highly accurate drafting within the predefined tool schema; and \textbf{(b) retrieval-augmented speculation}, which reuses historical generation patterns as high-quality drafts to speed up recurrent tool calls.}
\label{fig:toolspec}
\end{figure*}

As illustrated in Figure~\ref{fig:transition}, once the special token \texttt{<tool\_call>} is generated, the FSM transitions to the tool-name state $q_t$. In this state, drafting is executed via constrained decoding to generate candidate draft sequences. We formulate the proposed draft sequence $\tilde{\boldsymbol{y}}$ as the concatenation (denoted by $\oplus$) of the necessary structural elements:
\begin{equation}\label{eq:draft}
    \tilde{\boldsymbol{y}} = \tilde{\boldsymbol{y}}_{\text{prefix}} \oplus \tilde{\boldsymbol{y}}_{\text{tool\_name}} \oplus \tilde{\boldsymbol{y}}_{\text{suffix}}
\end{equation}
where $\tilde{\boldsymbol{y}}_{\text{prefix}}$ and $\tilde{\boldsymbol{y}}_{\text{suffix}}$ denote the corresponding schema prefix and suffix associated with $q_t$, and $\tilde{\boldsymbol{y}}_{\text{tool\_name}} \in \mathcal{C}_{\text{tools}}$ represents the candidate tool names. All potential drafts are subsequently verified in parallel by the target model $\mathcal{M}$.

Upon successful verification, the FSM transitions to the parameter-name state $q_p$, where $\mathcal{M}$ verifies the candidate parameter names associated with the selected tool in parallel. The FSM then transitions to the parameter-value state $q_v$. Because parameter values are typically unconstrained, \method eschews schema constraints at this stage and instead relies on a general SD algorithm to predict these variable fields.

Then, the FSM transitions between states according to schema-specific delimiters. For example, the delimiter \mybox{delimiter}{\texttt{'',}} terminates the parameter-value state $q_v$ and returns to $q_p$ for the next parameter; while the delimiter \mybox{delimiter}{\texttt{``\}\}''}} marks completion of the tool call. Consequently, \method seamlessly alternates between deterministic schema token filling for structural states (e.g., $q_t$ and $q_p$) and speculative generation for open-ended states (e.g., $q_v$ and $q_o$).

\subsection{Retrieval-augmented Speculation}
While the schema defines the structural skeleton of a tool call and supports local drafting for constrained fields such as tool names, variable fields, particularly parameter values, depend heavily on the input context and cannot be inferred from the schema alone. Motivated by the recurring patterns in tool invocations, \method introduces a retrieval-augmented drafting strategy to accelerate tool-calling generation at the sequence level.

\begin{table*}[!t]
\centering
\vspace{-1.0cm}
\small
\setlength{\tabcolsep}{1.4mm}
\resizebox{\linewidth}{!}{
\begin{tabular}{llcccccccccc}
\toprule
\multirow{2}{*}{\textbf{Models}} & \multirow{2}{*}{\textbf{Methods}}
& \multicolumn{3}{c}{\textbf{API-Bank}}  & \multicolumn{3}{c}{\textbf{ToolAlpaca}}   & \multicolumn{3}{c}{\textbf{BFCLv2}} & \multirow{2}{*}{\begin{tabular}[c]{@{}c@{}}\textbf{Overall} \\\textbf{Speedup}\end{tabular}} \\ \cmidrule(lr){3-5} \cmidrule(lr){6-8} \cmidrule(lr){9-11} 
& &\#MAT &Tokens/s & Speedup   &\#MAT &Tokens/s & Speedup &\#MAT &Tokens/s & Speedup  \\ \midrule
\multirow{5}{*}{\begin{tabular}[c]{@{}c@{}}Qwen2.5-7B\\-Instruct\end{tabular}} & Vanilla & 1.00 & 32.76 & 1.00$\times$  & 1.00 & 32.72 & 1.00$\times$  & 1.00 & 32.73 & 1.00$\times$  & 1.00$\times$ \\
& PLD &2.37 &64.56 &1.97$\times$ &2.10 &60.75 &1.86$\times$ &2.24 &62.52 &1.91$\times$ & 1.91$\times$ \\
& TR &3.09 &78.95 &2.41$\times$ &\underline{2.87} &\underline{71.28} &\underline{2.18$\times$} &2.89 &\underline{73.64} &\underline{2.25$\times$} & \underline{2.28$\times$} \\
& SAMD &\underline{3.60} &\underline{83.96} &\underline{2.56$\times$} &2.62 &63.57 &1.94$\times$ &\underline{2.92} &70.70 &2.16$\times$ & 2.22$\times$ \\
& \blue{\method} & \blue{\bf{4.90}} &\blue{\bf{129.40}} &\blue{\bf{3.95$\times$}} & \blue{\bf{4.55}} &\blue{\bf{113.54}} &\blue{\bf{3.47$\times$}} & \blue{\bf{4.73}} &\blue{\bf{117.17}} &\blue{\bf{3.58$\times$}} &\blue{\bf{3.67$\times$}} \\\midrule
\multirow{6}{*}{\begin{tabular}[c]{@{}c@{}}Qwen2.5-14B\\-Instruct\end{tabular}} & Vanilla & 1.00 & 19.50 & 1.00$\times$  & 1.00 & 19.02 & 1.00$\times$ & 1.00 & 19.31 & 1.00$\times$  &1.00$\times$  \\
& PLD &1.98 &35.30 &1.81$\times$ &2.00 &34.85 &1.83$\times$ &2.02 &35.34 &1.83$\times$ &1.82$\times$\\
& TR &3.01 &\underline{46.94} &\underline{2.41$\times$} &2.85 &\underline{41.97} &\underline{2.21$\times$} &2.99 &\underline{46.34} &\underline{2.40$\times$} &\underline{2.34$\times$}\\
& SAMD &2.93 &45.86 &2.35$\times$ &2.62 &38.48 &2.02$\times$ &2.78 &42.29 &2.19$\times$ &2.19$\times$\\
& \gray{Eagle-2$^*$} & \gray{\underline{3.87}} & \gray{46.80}  & \gray{2.40$\times$} & \gray{\underline{3.58}} & \gray{36.24} & \gray{1.91$\times$} &\gray{\underline{3.67}} & \gray{39.39} & \gray{2.04$\times$} &\gray{2.12$\times$} \\
& \blue{\method} & \blue{\bf{4.43}} &\blue{\bf{75.27}} &\blue{\bf{3.86$\times$}} & \blue{\bf{4.02}} &\blue{\bf{60.86}} &\blue{\bf{3.20$\times$}} & \blue{\bf{4.15}} &\blue{\bf{65.07}} &\blue{\bf{3.37$\times$}} &\blue{\bf{3.48$\times$}}\\\midrule
\multirow{6}{*}{\begin{tabular}[c]{@{}c@{}}LLaMA-3.1-8B\\-Instruct\end{tabular}} & Vanilla & 1.00 & 28.21 & 1.00$\times$ & 1.00 & 27.69 & 1.00$\times$ & 1.00 & 27.83 & 1.00$\times$ &1.00$\times$ \\
& PLD & 2.35 & 60.72  & 2.15$\times$ & 2.27 & 58.28 & 2.10$\times$ &2.30 & 58.72 &2.11$\times$ &2.12$\times$ \\
& TR & \underline{3.34} & \underline{71.10}  & \underline{2.52$\times$} & \underline{3.32} & \underline{67.69} & \underline{2.44$\times$} &\underline{3.29} &\underline{66.79} &\underline{2.40$\times$} &\underline{2.45$\times$} \\
& SAMD  & 3.20 & 69.68 & 2.47$\times$ & 3.12 & 65.82 & 2.38$\times$ &3.14 &66.24 & 2.38$\times$ &2.41$\times$ \\
& \gray{Eagle-3$^*$} & \gray{\underline{3.34}} & \gray{47.24}  & \gray{1.67$\times$} & \gray{2.93} & \gray{40.37} & \gray{1.46$\times$} &\gray{3.17} & \gray{42.86} & \gray{1.54$\times$} &\gray{1.56$\times$} \\
& \blue{\method} &\blue{\bf{5.80}} &\blue{\bf{125.59}} &\blue{\bf{4.45$\times$}} &\blue{\bf{5.26}} &\blue{\bf{112.26}} &\blue{\bf{4.05$\times$}} &\blue{\bf{5.31}} &\blue{\bf{113.55}} &\blue{\bf{4.08$\times$}} &\blue{\bf{4.19$\times$}} \\\midrule
\multirow{5}{*}{\begin{tabular}[c]{@{}c@{}}LLaMA-3.2-3B\\-Instruct\end{tabular}} & Vanilla & 1.00 & 32.60 & 1.00$\times$  & 1.00 & 33.20 & 1.00$\times$  & 1.00 & 32.91 & 1.00$\times$  & 1.00$\times$  \\
& PLD &2.19 &69.81 &2.14$\times$ &2.06 &61.41 &1.85$\times$ &2.11 &64.83 &1.97$\times$ &1.99$\times$\\
& TR &\underline{3.11} &\underline{75.50} &\underline{2.32$\times$} &\underline{3.03} &\underline{73.45} &\underline{2.31$\times$} &\underline{3.06} &\underline{76.35} &\underline{2.32$\times$} &\underline{2.32$\times$}\\
& SAMD &2.99 &72.79 &2.18$\times$ &2.68 &71.84 &2.16$\times$ &2.75 &71.09 &2.16$\times$ &2.17$\times$\\
& \blue{\method} & \blue{\bf{5.20}} &\blue{\bf{133.30}} &\blue{\bf{4.09$\times$}} & \blue{\bf{4.52}} &\blue{\bf{112.22}} &\blue{\bf{3.38$\times$}} & \blue{\bf{4.78}} &\blue{\bf{115.84}} &\blue{\bf{3.52$\times$}} &\blue{\bf{3.66$\times$}}\\
\bottomrule
\end{tabular}}
\caption{Comparison of inference efficiency between \method and prior plug-and-play methods. We report the mean number of accepted tokens (\#MAT), average decoding speed (tokens/s), and wall-clock speedup relative to vanilla autoregressive decoding. The best results are shown in \textbf{boldface}, and the second-best results are \underline{underlined}. \gray{$^*$The Eagle series, a training-required SD method, is included for reference only.}}
\label{tab:main-exp}
\end{table*}

We maintain a datastore $\mathcal{H}=\{(\boldsymbol{h}_j, \boldsymbol{y}_j)\}_{j=1}^{|\mathcal{H}|}$ consisting of prior successful tool calls, where $\boldsymbol{h}_j \in \mathbb{R}^d$ denotes the hidden representation of the input question $\boldsymbol{x}_j$ at its final token, $d$ is the hidden dimension, and $\boldsymbol{y}_j$ is the corresponding tool-call output. Given a query $\boldsymbol{x}_i$, we retrieve the top-$k$ most similar historical tool calls:

\begin{equation}
    \mathcal{I}_k = \operatorname{Top-K}_{j \in \{1 \ldots |\mathcal{H}|\}} \text{sim}(\boldsymbol{h}_i, \boldsymbol{h}_j)
\end{equation}
where $\text{sim}(\cdot)$ denotes cosine similarity and $\mathcal{I}_k$ the indices of the retrieved entries. Given the partially generated output $\boldsymbol{y}_{<t}$, we align it with each retrieved tool call $\boldsymbol{y}^{\prime}$ by matching the suffix:
\begin{equation}
\boldsymbol{y}_{t-L:t-1}=\boldsymbol{y}^{\prime}_{m-L:m-1}
\end{equation}
where $L$ denotes the suffix length and $m$ indicates the corresponding match position in $\boldsymbol{y}^{\prime}$. When a suffix match is found, \method extracts the next $n$ tokens $\boldsymbol{y}^{\prime}_{m:m+n-1}$, using these continuations as drafts. As shown in Figure~\ref{fig:toolspec}, if the current suffix matches a historical invocation containing \mybox{delimiter}{...``parameters''}, the subsequent tokens (e.g., \mybox{delimiter}{:\{``status'': ``Forgot Password''...}) are proposed as draft candidates. Intuitively, once the tool name and key–value pairs are aligned, the remaining structure of the tool call often follows recurring patterns, making $\boldsymbol{y}'_{m:m+n-1}$ high-quality drafts for subsequent value spans.

\section{Experiments}

\subsection{Experimental Setup}
\label{sec:exp_setup}

\paragraph{Models and Datasets} 
We primarily evaluate our method using the Qwen2.5-Instruct~\citep{Qwen2.5} series, ToolLLaMA~\citep{Qin:2024toolllm} series, LLaMA-3.1-8B-Instruct, and LLaMA-3.2-3B-Instruct~\citep{llama3}. The evaluation covers four widely used tool-use benchmarks: API-Bank~\citep{apibank}, ToolAlpaca~\cite{ToolAlpaca}, BFCLv2~\cite{BFCL}, and ToolBench~\cite{Qin:2024toolllm}. Additional details of these datasets are provided in the Appendix~\ref{appendix:dataset_details}.

\paragraph{Implementation Details} 
The retrieval parameter $k$ is set to 10. For suffix matching, we follow PLD~\citep{saxena2023pld} and consider suffix lengths $L \in \{5,6,7\}$. The continuations are extracted using predefined lengths $n \in \{32, 16, 8, 8\}$. We adopt Token Recycling (TR)~\citep{luo:2025tokenrecycle} as the general speculative strategy in \method. Following prior work, we adopt speculative sampling~\citep{leviathan:2023spec} as our acceptance strategy, with a batch size of 1. Additional implementation details are provided in the Appendix~\ref{appendix:impl_details}.

\paragraph{Baselines} 
We primarily compare \method to three existing \textit{plug-and-play} methods: PLD~\citep{saxena2023pld}, TR~\citep{luo:2025tokenrecycle}, and SAM-Decoding (SAMD)~\citep{hu2025:samd}. For SAMD, we construct the static SAM from the input context and adopt TR as its decoding strategy. For comparison, we also include the advanced training-required methods Eagle-2~\citep{li2024:eagle2} and Eagle-3~\citep{li2025:eagle3} in our main results.

\begin{figure}[t]
\centering
\includegraphics[width=0.95\columnwidth]{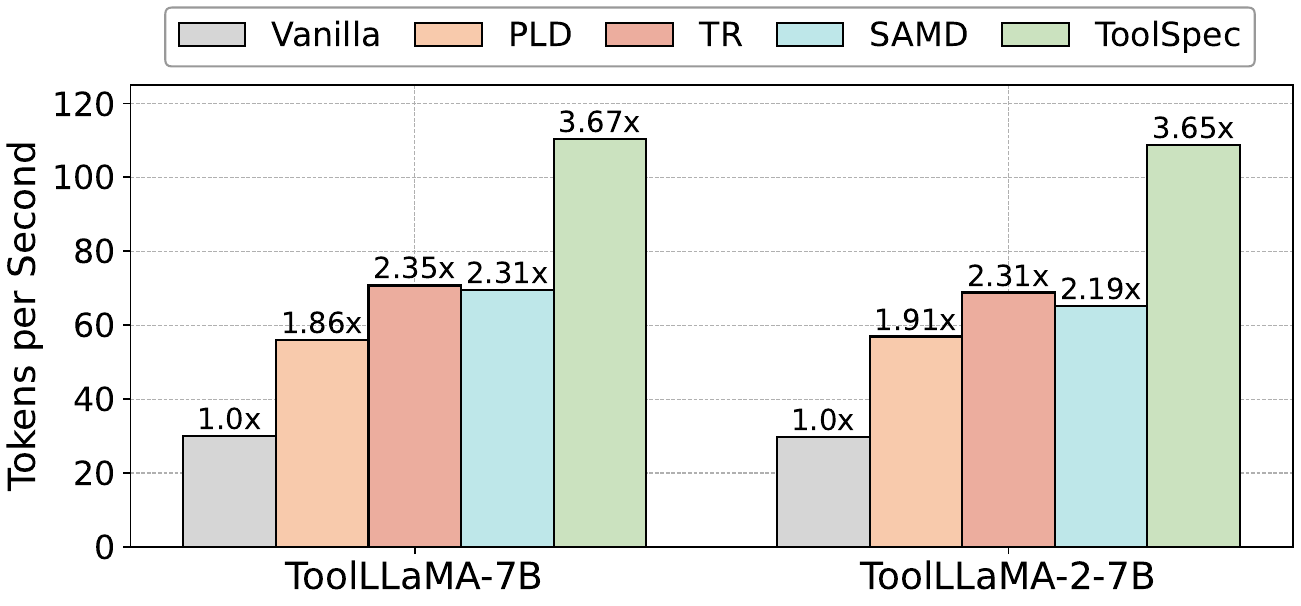}
\caption{Speedup comparison between \method and prior plug-and-play methods on ToolBench.}
\vspace{-0.5cm}
\label{fig:toolllama}
\end{figure}

\paragraph{Evaluation Metrics} 
We evaluate \method using three widely adopted metrics: the mean number of accepted tokens (\#MAT), throughput (tokens/s), and wall-clock speedup relative to vanilla autoregressive decoding. From a theoretical perspective, \method preserves the output distribution of the target LLM, thereby obviating the need for extensive evaluation of generation quality. Nevertheless, we report performance on API-Bank and ToolAlpaca in Appendix~\ref{appendix:performance} for reference.

\subsection{Main Results} 
Table~\ref{tab:main-exp} compares the inference efficiency of \method with prior plug-and-play methods on widely used tool-use tasks. The results reveal the following key findings: (1) \method consistently outperforms existing approaches, achieving speedups of $3.5\times$$\sim$$4.2\times$ over standard autoregressive decoding across a range of models and tasks. Compared with the previous best method, it yields up to a $\bf{71\%}$ relative improvement in speedup (from $2.45\times$ to $4.19\times$). (2) The efficiency gains of \method are primarily attributed to a high number of accepted tokens (denoted as \#MAT), ranging from $4.02$ to $5.80$ across different settings. Unlike model-based SD methods such as Eagle-3, \method does not require an additional drafting module and introduces only minimal drafting overhead (approximately $6\%$, as shown in Figure~\ref{fig:profile}). This low overhead enables the increased \#MAT to translate more effectively into end-to-end speedup.

\paragraph{Discussion on Eagle Performance}
We observe that the \#MAT achieved by the Eagle series is lower than that of prior plug-and-play methods, such as TR, as shown in Table~\ref{tab:main-exp}. We hypothesize that this gap arises because Eagle models are not trained on tool-calling formats, making such inputs out-of-distribution for these models. In contrast, as discussed in Section~\ref{sec:pattern_analysis}, tool-calling traces exhibit structural regularities and recurring invocation patterns. These properties favor SD methods such as TR, which leverage historical traces to construct draft sequences. Building on this insight, \method exploits the schema-constrained nature of tool-calling tasks, enabling more accurate draft generation and achieving higher overall efficiency.

\paragraph{Results on ToolLLaMA}
In addition to general-purpose LLMs, we further evaluate \method on two variants of ToolLLaMA, a model specifically trained for tool-use tasks. Figure~\ref{fig:toolllama} presents a comparison of inference efficiency between \method and prior plug-and-play SD methods on ToolBench. The results show that \method achieves a speedup of approximately $3.7\times$ over standard autoregressive decoding, demonstrating its effectiveness on specialized tool-use LLMs.
\section{Analysis}
\label{sec:analysis}

\begin{figure}[t]
\centering
\vspace{-0.5cm}
\includegraphics[width=0.95\columnwidth]{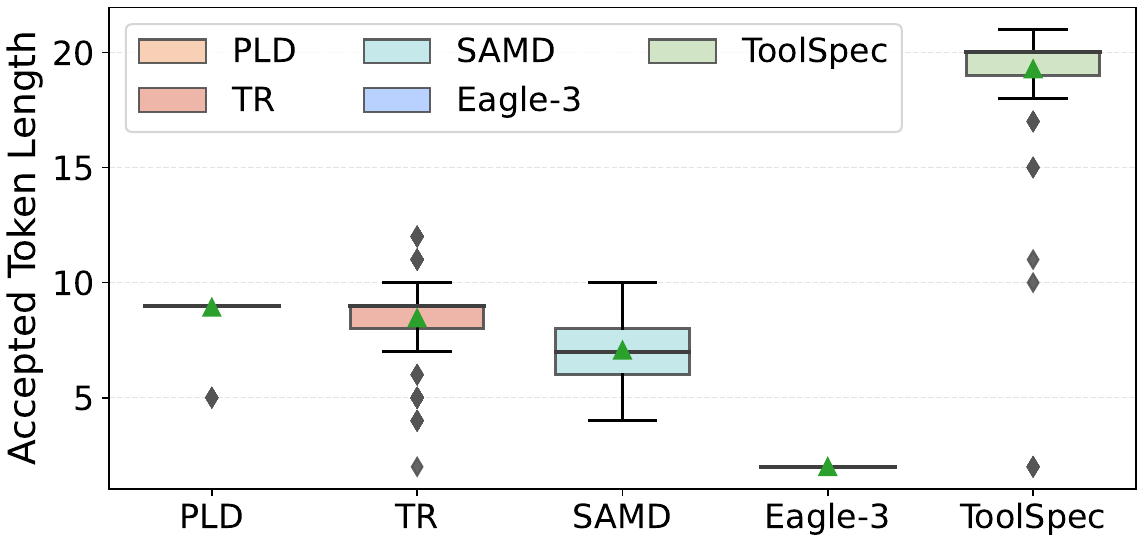}
\caption{Distribution of accepted token lengths in the first two decoding steps using LLaMA-3.1-8B-Instruct on API-Bank. Green triangles represent mean values, and black diamonds denote outliers.}
\label{fig:first2token}
\end{figure}

This section presents a comprehensive analysis of \method. Specifically, we evaluate its effectiveness (\S\ref{sec:effectiveness}), examine the impact of adherence to the tool-calling format (\S\ref{sec:adherence_impact}), and analyze the time allocation across different modules (\S\ref{sec:profile}).

\subsection{Effectiveness of \method}
\label{sec:effectiveness}

\begin{figure}[t]
\centering
\vspace{-0.5cm}
\includegraphics[width=0.95\columnwidth]{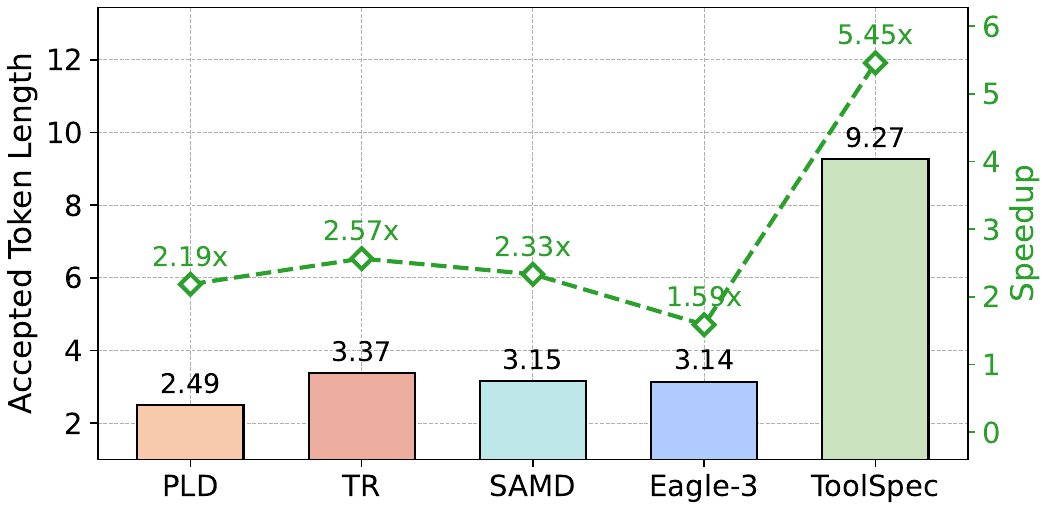}
\caption{Mean number of accepted tokens (\#MAT) and speedups for repeated tool calls. Results are obtained using LLaMA-3.1-8B-Instruct on API-Bank.}
\label{fig:repeat_call}
\end{figure}

\paragraph{Statistical Analysis}
We first conduct a statistical evaluation to assess the effectiveness of the proposed modules, including: \textbf{1) Schema-aware Drafting.} Figure~\ref{fig:first2token} reports accepted token lengths in the first two decoding steps, corresponding to the tool-name state $q_t$ and the first parameter-name state $q_p$ in \method. As shown, schema-aware drafting enables \method to generate an average of $19.3$ tokens across these steps. In contrast, prior approaches such as TR produce an average of $8.5$ tokens, while Eagle-3 yields only $2$ tokens on average, equivalent to standard autoregressive decoding. These results demonstrate the clear advantage of schema-aware drafting for tool-calling generation. \textbf{2) Retrieval-augmented Speculation.} Figure~\ref{fig:repeat_call} presents \#MAT for repeated tool calls. The results indicate that \method consistently outperforms prior methods, achieving an average of $9.3$ accepted tokens per decoding step and approximately $5.5\times$ speedup. This improvement highlights the effectiveness of the proposed retrieval strategy in handling repetitive tool invocations.

\begin{figure}[t]
\centering
\includegraphics[width=0.95\columnwidth]{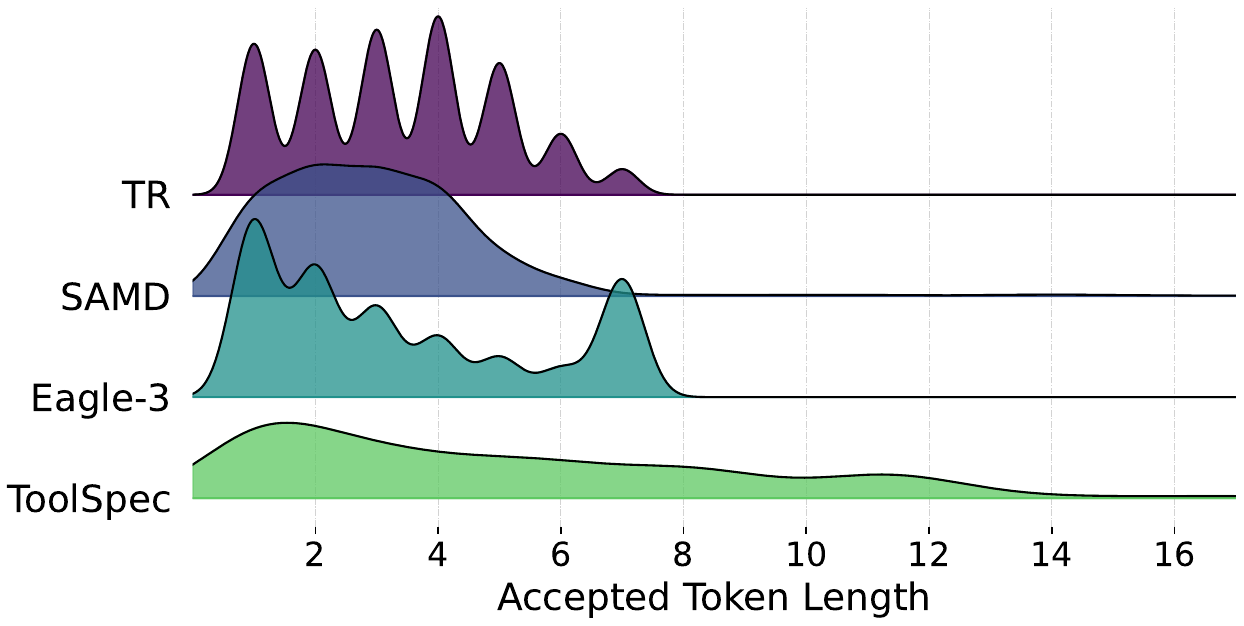}
\caption{Distribution of accepted token lengths using LLaMA-3.1-8B-Instruct on API-Bank. Values exceeding 17 are truncated for clearer visualization.}
\vspace{-0.3cm}
\label{fig:ridge}
\end{figure}

\paragraph{Acceptance Distribution}
Figure~\ref{fig:ridge} presents the distribution of accepted token lengths for \method in comparison with prior methods. While existing approaches typically yield accepted token lengths of fewer than $8$ tokens, \method shifts the distribution toward longer accepted lengths, with $22\%$ of decoding steps exceeding $8$ tokens. Furthermore, \method exhibits a pronounced long-tail distribution: nearly $4\%$ of decoding steps attain accepted token lengths between $17$ and $33$. This long-tail behavior stems from retrieval-augmented speculation, which reuses historical tool-call trajectories to produce longer and more accurate drafts. These consistently higher accepted token lengths contribute to the overall speedup achieved by \method.

\begin{table}[t]
\centering
\vspace{-0.5cm}
\small
\begin{tabular}{@{}l|ccc@{}}
\toprule
\bf Methods &\#MAT &Tokens/s & Speedup \\\midrule
Vanilla & 1.00 & 28.21 & 1.00$\times$ \\ \midrule
\method & \bf{5.80} & \bf{125.59} & \bf{4.45$\times$}  \\
\quad \textit{w/o SAD} & 5.58 & 118.99 & 4.22$\times$ \\
\quad \textit{w/o RAS} & 4.29 & 94.53 & 3.35$\times$ \\
\quad \textit{w/o both} & 3.62 & 76.00 & 2.69$\times$  \\
\bottomrule
\end{tabular}
\caption{Ablation study of \method modules. ``\textit{SAD}'' and ``\textit{RAS}'' denote schema-aware drafting and retrieval-augmented speculation, respectively. Results are obtained with LLaMA-3.1-8B-Instruct on API-Bank.}
\label{tab:ablation}
\end{table}

\paragraph{Ablation Study}
Table~\ref{tab:ablation} illustrates an ablation study of \method and its proposed components. The results indicate that both schema-aware drafting and retrieval-augmented speculation are critical to the efficiency of \method. When these components are removed, similar to PLD, \method relies on suffix matching within the input context, resulting in slightly higher speedup than TR.

\begin{figure}[t]
\centering
\includegraphics[width=0.95\columnwidth]{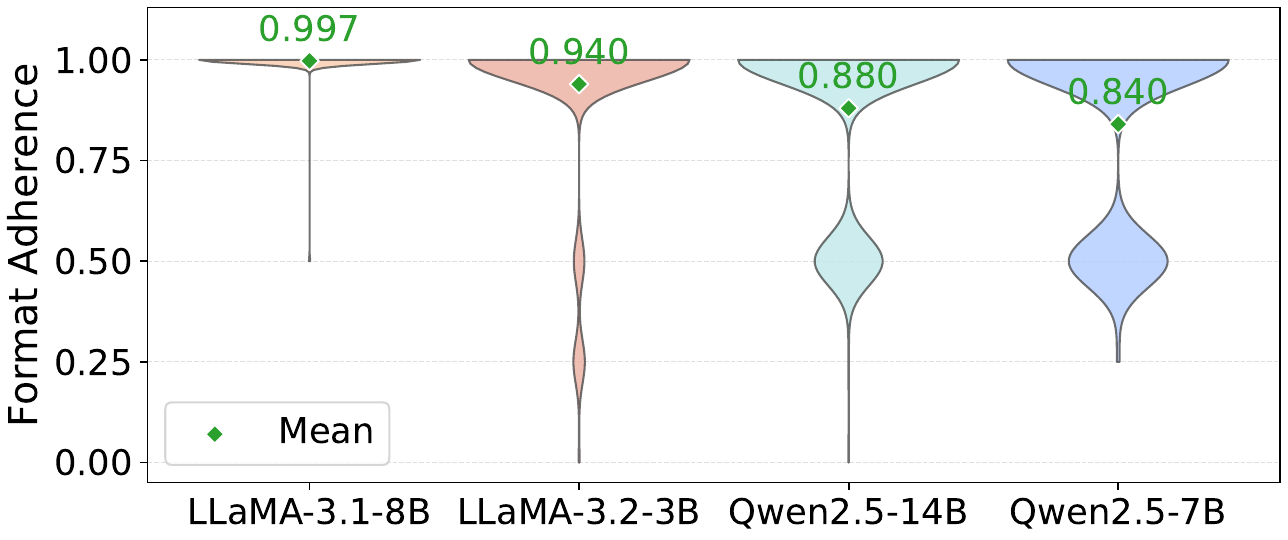}
\caption{Format adherence of different models in tool-calling generation. Results are obtained on API-Bank.}
\label{fig:adherence}
\end{figure}

\subsection{Impact of Format Adherence}
\label{sec:adherence_impact}
Beyond the analysis in Section~\ref{sec:pattern_analysis}, we further evaluate how strictly tool-use LLMs follow the required schema during tool-calling generation. We define strict format adherence using two criteria: 1) the outputs must be correctly parsed as valid JSON; and 2) any extraneous natural language, Markdown fences, XML-like tags, or malformed JSON structures are treated as violations. The results reveal a strong positive correlation between format adherence and the speedup achieved by \method, with a Pearson correlation coefficient of approximately $0.90$. Notably, LLaMA-3.1-8B-Instruct achieves the highest adherence score of $0.997$, corresponding to a $4.45\times$ speedup on API-Bank.

\begin{figure}[t]
\centering
\vspace{-0.5cm}
\includegraphics[width=0.85\columnwidth]{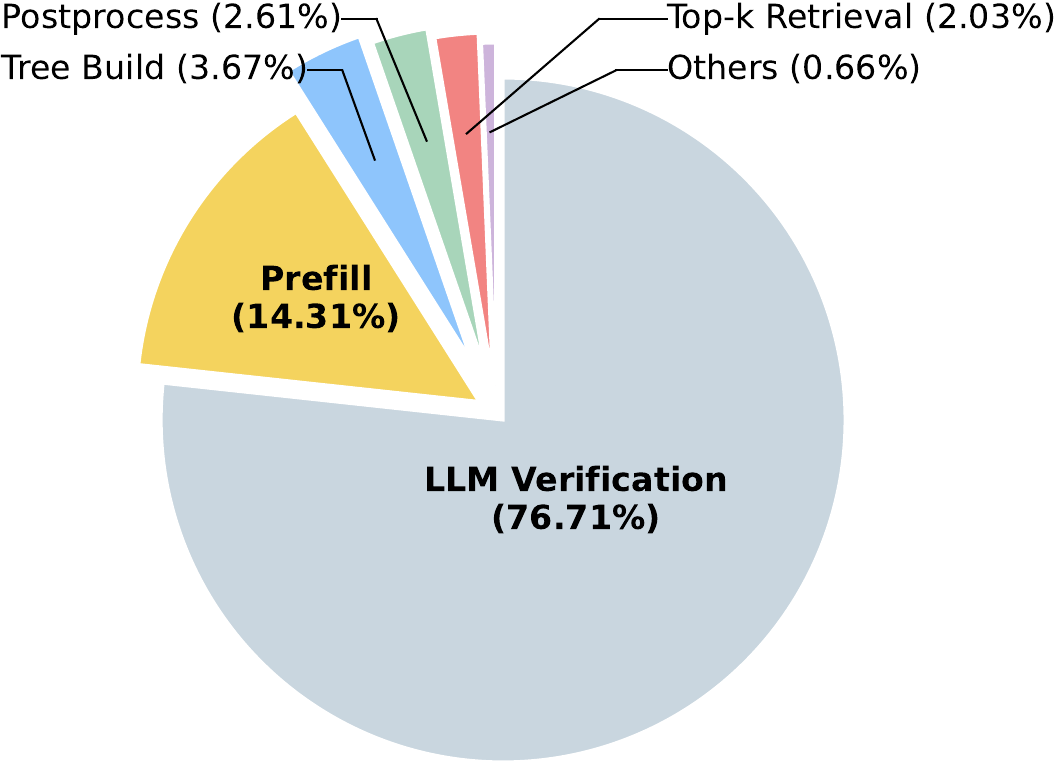}
\caption{Time allocation for each operation when LLMs respond to a query. Results are obtained using LLaMA-3.1-8B-Instruct on API-Bank.}
\label{fig:profile}
\end{figure}

\subsection{Time Allocation}
\label{sec:profile}

Figure~\ref{fig:profile} presents the time allocation across the primary latency-intensive components of \method. As shown, the decoding stage of the target model dominates overall latency, accounting for $91.0\%$ of the total runtime. In contrast, the additional operations introduced by \method, including top-$k$ retrieval and draft tree construction, incur only modest overhead, contributing just $5.7\%$ of the total latency. Despite this minimal overhead, \method achieves an approximately $4\times$ speedup in tool-calling generation, demonstrating its effectiveness in reducing end-to-end latency.

\section{Conclusion}
\label{sec:conclusion}
This work presents \method, a schema-aware, retrieval-augmented speculative decoding approach for accelerating tool-calling generation. It requires no additional training or auxiliary models, enabling seamless integration into existing LLM workflows. Extensive experiments across various LLMs and tool-use benchmarks demonstrate that \method achieves a $3.5\times$$\sim$$4.2\times$ speedup while preserving output distributions. Further analysis highlights the effectiveness of the proposed components and the impact of format adherence on the efficiency of \method. We hope this work advances efficient tool use in LLMs and provides practical insights into scalable tool-calling generation.
\section*{Limitations}
\label{sec:limitation} 
Due to computational constraints, our experiments do not include larger-scale language models such as Qwen2.5-72B-Instruct~\citep{Qwen2.5} and LLaMA-3.1-70B-Instruct~\cite{llama3}. Nevertheless, we expect \method to generalize well to these models and achieve further improvements in tool-calling generation efficiency. We include experimental results of Qwen2.5-32B-Instruct in Appendix~\ref{appendix:qwen32B}. Additionally, our approach assumes strong adherence to predefined tool schemas and the existence of recurring invocation patterns, which are consistent with real-world application scenarios characterized by frequent tool usage. As discussed in Sections~\ref{sec:preliminary} and \ref{sec:adherence_impact}, we anticipate that the effectiveness of \method will continue to improve as LLM capabilities advance.

\section*{Ethics Statement}
\label{sec:ethics} 
The datasets used in our experiments are publicly released and labeled through interaction with humans in English. In this process, user privacy is protected, and no personal information is contained in the dataset. The scientific artifacts that we used are available for research with permissive licenses. The use of these artifacts in this paper is consistent with their intended use. Therefore, we believe that our research work meets ACL's ethical standards. 

\bibliography{custom}

@InProceedings{leviathan:2023spec,
  title = 	 {Fast Inference from Transformers via Speculative Decoding},
  author =       {Leviathan, Yaniv and Kalman, Matan and Matias, Yossi},
  booktitle = 	 {Proceedings of the 40th International Conference on Machine Learning},
  pages = 	 {19274--19286},
  year = 	 {2023},
  editor = 	 {Krause, Andreas and Brunskill, Emma and Cho, Kyunghyun and Engelhardt, Barbara and Sabato, Sivan and Scarlett, Jonathan},
  volume = 	 {202},
  series = 	 {Proceedings of Machine Learning Research},
  month = 	 {23--29 Jul},
  publisher =    {PMLR},
  url = 	 {https://proceedings.mlr.press/v202/leviathan23a.html},
}

@article{o1,
  author       = {OpenAI},
  title        = {OpenAI o1 System Card},
  journal      = {CoRR},
  volume       = {abs/2412.16720},
  year         = {2024},
  url          = {https://doi.org/10.48550/arXiv.2412.16720},
  doi          = {10.48550/ARXIV.2412.16720},
  eprinttype   = {arXiv},
  eprint       = {2412.16720},
  bibsource    = {dblp computer science bibliography, https://dblp.org}
}

@misc{saxena2023pld,
    title = {Prompt Lookup Decoding},
    author = {Apoorv Saxena},
    year = {2023},
    month = {November},
    url = {https://github.com/apoorvumang/prompt-lookup-decoding/}
}

@article{Qwen3,
  author       = {An Yang and
                  Anfeng Li and
                  Baosong Yang and
                  Beichen Zhang and
                  Binyuan Hui and
                  Bo Zheng and
                  Bowen Yu and
                  Chang Gao and
                  Chengen Huang and
                  Chenxu Lv and
                  Chujie Zheng and
                  Dayiheng Liu and
                  Fan Zhou and
                  Fei Huang and
                  Feng Hu and
                  Hao Ge and
                  Haoran Wei and
                  Huan Lin and
                  Jialong Tang and
                  Jian Yang and
                  Jianhong Tu and
                  Jianwei Zhang and
                  Jian Yang and
                  Jiaxi Yang and
                  Jingren Zhou and
                  Jingren Zhou and
                  Junyang Lin and
                  Kai Dang and
                  Keqin Bao and
                  Kexin Yang and
                  Le Yu and
                  Lianghao Deng and
                  Mei Li and
                  Mingfeng Xue and
                  Mingze Li and
                  Pei Zhang and
                  Peng Wang and
                  Qin Zhu and
                  Rui Men and
                  Ruize Gao and
                  Shixuan Liu and
                  Shuang Luo and
                  Tianhao Li and
                  Tianyi Tang and
                  Wenbiao Yin and
                  Xingzhang Ren and
                  Xinyu Wang and
                  Xinyu Zhang and
                  Xuancheng Ren and
                  Yang Fan and
                  Yang Su and
                  Yichang Zhang and
                  Yinger Zhang and
                  Yu Wan and
                  Yuqiong Liu and
                  Zekun Wang and
                  Zeyu Cui and
                  Zhenru Zhang and
                  Zhipeng Zhou and
                  Zihan Qiu},
  title        = {Qwen3 Technical Report},
  journal      = {CoRR},
  volume       = {abs/2505.09388},
  year         = {2025},
  url          = {https://doi.org/10.48550/arXiv.2505.09388},
  doi          = {10.48550/ARXIV.2505.09388},
  eprinttype    = {arXiv},
  eprint       = {2505.09388},
  bibsource    = {dblp computer science bibliography, https://dblp.org}
}

@InProceedings{PAL,
  title = 	 {{PAL}: Program-aided Language Models},
  author =       {Gao, Luyu and Madaan, Aman and Zhou, Shuyan and Alon, Uri and Liu, Pengfei and Yang, Yiming and Callan, Jamie and Neubig, Graham},
  booktitle = 	 {Proceedings of the 40th International Conference on Machine Learning},
  pages = 	 {10764--10799},
  year = 	 {2023},
  editor = 	 {Krause, Andreas and Brunskill, Emma and Cho, Kyunghyun and Engelhardt, Barbara and Sabato, Sivan and Scarlett, Jonathan},
  volume = 	 {202},
  series = 	 {Proceedings of Machine Learning Research},
  month = 	 {23--29 Jul},
  publisher =    {PMLR},
  url = 	 {https://proceedings.mlr.press/v202/gao23f.html},
}

@inproceedings{SWIFT,
  author       = {Heming Xia and
                  Yongqi Li and
                  Jun Zhang and
                  Cunxiao Du and
                  Wenjie Li},
  title        = {{SWIFT:} On-the-Fly Self-Speculative Decoding for {LLM} Inference
                  Acceleration},
  booktitle    = {The Thirteenth International Conference on Learning Representations,
                  {ICLR} 2025, Singapore, April 24-28, 2025},
  publisher    = {OpenReview.net},
  year         = {2025},
  url          = {https://openreview.net/forum?id=EKJhH5D5wA},
  bibsource    = {dblp computer science bibliography, https://dblp.org}
}

@inproceedings{Qin:2024toolllm,
  author       = {Yujia Qin and
                  Shihao Liang and
                  Yining Ye and
                  Kunlun Zhu and
                  Lan Yan and
                  Yaxi Lu and
                  Yankai Lin and
                  Xin Cong and
                  Xiangru Tang and
                  Bill Qian and
                  Sihan Zhao and
                  Lauren Hong and
                  Runchu Tian and
                  Ruobing Xie and
                  Jie Zhou and
                  Mark Gerstein and
                  Dahai Li and
                  Zhiyuan Liu and
                  Maosong Sun},
  title        = {ToolLLM: Facilitating Large Language Models to Master 16000+ Real-world
                  APIs},
  booktitle    = {The Twelfth International Conference on Learning Representations,
                  {ICLR} 2024, Vienna, Austria, May 7-11, 2024},
  publisher    = {OpenReview.net},
  year         = {2024},
  url          = {https://openreview.net/forum?id=dHng2O0Jjr},
  bibsource    = {dblp computer science bibliography, https://dblp.org}
}

@inproceedings{BFCL,
  author       = {Shishir G. Patil and
                  Huanzhi Mao and
                  Fanjia Yan and
                  Charlie Cheng{-}Jie Ji and
                  Vishnu Suresh and
                  Ion Stoica and
                  Joseph E. Gonzalez},
  editor       = {Aarti Singh and
                  Maryam Fazel and
                  Daniel Hsu and
                  Simon Lacoste{-}Julien and
                  Felix Berkenkamp and
                  Tegan Maharaj and
                  Kiri Wagstaff and
                  Jerry Zhu},
  title        = {The Berkeley Function Calling Leaderboard {(BFCL):} From Tool Use
                  to Agentic Evaluation of Large Language Models},
  booktitle    = {Forty-second International Conference on Machine Learning, {ICML}
                  2025, Vancouver, BC, Canada, July 13-19, 2025},
  series       = {Proceedings of Machine Learning Research},
  publisher    = {{PMLR} / OpenReview.net},
  year         = {2025},
  url          = {https://proceedings.mlr.press/v267/patil25a.html},
  bibsource    = {dblp computer science bibliography, https://dblp.org}
}

@article{ToolAlpaca,
  author       = {Qiaoyu Tang and
                  Ziliang Deng and
                  Hongyu Lin and
                  Xianpei Han and
                  Qiao Liang and
                  Le Sun},
  title        = {ToolAlpaca: Generalized Tool Learning for Language Models with 3000
                  Simulated Cases},
  journal      = {CoRR},
  volume       = {abs/2306.05301},
  year         = {2023},
  url          = {https://doi.org/10.48550/arXiv.2306.05301},
  doi          = {10.48550/ARXIV.2306.05301},
  eprinttype   = {arXiv},
  eprint       = {2306.05301},
  bibsource    = {dblp computer science bibliography, https://dblp.org}
}

@inproceedings{yao2025taubench,
title={\{\${\textbackslash}tau\$\}-bench: A Benchmark for {\textbackslash}underline\{T\}ool-{\textbackslash}underline\{A\}gent-{\textbackslash}underline\{U\}ser Interaction in Real-World Domains},
author={Shunyu Yao and Noah Shinn and Pedram Razavi and Karthik R Narasimhan},
booktitle={The Thirteenth International Conference on Learning Representations},
year={2025},
url={https://openreview.net/forum?id=roNSXZpUDN}
}

@inproceedings{Schick:2023toolformer,
  author       = {Timo Schick and
                  Jane Dwivedi{-}Yu and
                  Roberto Dess{\`{\i}} and
                  Roberta Raileanu and
                  Maria Lomeli and
                  Eric Hambro and
                  Luke Zettlemoyer and
                  Nicola Cancedda and
                  Thomas Scialom},
  editor       = {Alice Oh and
                  Tristan Naumann and
                  Amir Globerson and
                  Kate Saenko and
                  Moritz Hardt and
                  Sergey Levine},
  title        = {Toolformer: Language Models Can Teach Themselves to Use Tools},
  booktitle    = {Advances in Neural Information Processing Systems 36: Annual Conference
                  on Neural Information Processing Systems 2023, NeurIPS 2023, New Orleans,
                  LA, USA, December 10 - 16, 2023},
  year         = {2023},
  url          = {http://papers.nips.cc/paper\_files/paper/2023/hash/d842425e4bf79ba039352da0f658a906-Abstract-Conference.html},
  bibsource    = {dblp computer science bibliography, https://dblp.org}
}

@inproceedings{Gou:2024Tora,
  author       = {Zhibin Gou and
                  Zhihong Shao and
                  Yeyun Gong and
                  Yelong Shen and
                  Yujiu Yang and
                  Minlie Huang and
                  Nan Duan and
                  Weizhu Chen},
  title        = {ToRA: {A} Tool-Integrated Reasoning Agent for Mathematical Problem
                  Solving},
  booktitle    = {The Twelfth International Conference on Learning Representations,
                  {ICLR} 2024, Vienna, Austria, May 7-11, 2024},
  publisher    = {OpenReview.net},
  year         = {2024},
  url          = {https://openreview.net/forum?id=Ep0TtjVoap},
  bibsource    = {dblp computer science bibliography, https://dblp.org}
}

@article{Qin2024:ToolSurvey,
author = {Qin, Yujia and Hu, Shengding and Lin, Yankai and Chen, Weize and Ding, Ning and Cui, Ganqu and Zeng, Zheni and Zhou, Xuanhe and Huang, Yufei and Xiao, Chaojun and Han, Chi and Fung, Yi Ren and Su, Yusheng and Wang, Huadong and Qian, Cheng and Tian, Runchu and Zhu, Kunlun and Liang, Shihao and Shen, Xingyu and Xu, Bokai and Zhang, Zhen and Ye, Yining and Li, Bowen and Tang, Ziwei and Yi, Jing and Zhu, Yuzhang and Dai, Zhenning and Yan, Lan and Cong, Xin and Lu, Yaxi and Zhao, Weilin and Huang, Yuxiang and Yan, Junxi and Han, Xu and Sun, Xian and Li, Dahai and Phang, Jason and Yang, Cheng and Wu, Tongshuang and Ji, Heng and Li, Guoliang and Liu, Zhiyuan and Sun, Maosong},
title = {Tool Learning with Foundation Models},
year = {2024},
issue_date = {April 2025},
publisher = {Association for Computing Machinery},
address = {New York, NY, USA},
volume = {57},
number = {4},
issn = {0360-0300},
url = {https://doi.org/10.1145/3704435},
doi = {10.1145/3704435},
journal = {ACM Comput. Surv.},
month = dec,
articleno = {101},
numpages = {40}
}

@inproceedings{deepresearcher,
    title = "{D}eep{R}esearcher: Scaling Deep Research via Reinforcement Learning in Real-world Environments",
    author = "Zheng, Yuxiang  and
      Fu, Dayuan  and
      Hu, Xiangkun  and
      Cai, Xiaojie  and
      Ye, Lyumanshan  and
      Lu, Pengrui  and
      Liu, Pengfei",
    editor = "Christodoulopoulos, Christos  and
      Chakraborty, Tanmoy  and
      Rose, Carolyn  and
      Peng, Violet",
    booktitle = "Proceedings of the 2025 Conference on Empirical Methods in Natural Language Processing",
    month = nov,
    year = "2025",
    address = "Suzhou, China",
    publisher = "Association for Computational Linguistics",
    url = "https://aclanthology.org/2025.emnlp-main.22/",
    doi = "10.18653/v1/2025.emnlp-main.22",
    pages = "414--431",
    ISBN = "979-8-89176-332-6",
}

@inproceedings{hu2025:samd,
    title = "{SAM} Decoding: Speculative Decoding via Suffix Automaton",
    author = "Hu, Yuxuan  and
      Wang, Ke  and
      Zhang, Xiaokang  and
      Zhang, Fanjin  and
      Li, Cuiping  and
      Chen, Hong  and
      Zhang, Jing",
    editor = "Che, Wanxiang  and
      Nabende, Joyce  and
      Shutova, Ekaterina  and
      Pilehvar, Mohammad Taher",
    booktitle = "Proceedings of the 63rd Annual Meeting of the Association for Computational Linguistics (Volume 1: Long Papers)",
    month = jul,
    year = "2025",
    address = "Vienna, Austria",
    publisher = "Association for Computational Linguistics",
    url = "https://aclanthology.org/2025.acl-long.595/",
    doi = "10.18653/v1/2025.acl-long.595",
    pages = "12187--12204",
    ISBN = "979-8-89176-251-0",
}

@inproceedings{eagle,
  author       = {Yuhui Li and
                  Fangyun Wei and
                  Chao Zhang and
                  Hongyang Zhang},
  title        = {{EAGLE:} Speculative Sampling Requires Rethinking Feature Uncertainty},
  booktitle    = {Forty-first International Conference on Machine Learning, {ICML} 2024,
                  Vienna, Austria, July 21-27, 2024},
  publisher    = {OpenReview.net},
  year         = {2024},
  url          = {https://openreview.net/forum?id=1NdN7eXyb4},
  bibsource    = {dblp computer science bibliography, https://dblp.org}
}

@inproceedings{li2024:eagle2,
    title = "{EAGLE}-2: Faster Inference of Language Models with Dynamic Draft Trees",
    author = "Li, Yuhui  and
      Wei, Fangyun  and
      Zhang, Chao  and
      Zhang, Hongyang",
    editor = "Al-Onaizan, Yaser  and
      Bansal, Mohit  and
      Chen, Yun-Nung",
    booktitle = "Proceedings of the 2024 Conference on Empirical Methods in Natural Language Processing",
    month = nov,
    year = "2024",
    address = "Miami, Florida, USA",
    publisher = "Association for Computational Linguistics",
    url = "https://aclanthology.org/2024.emnlp-main.422/",
    doi = "10.18653/v1/2024.emnlp-main.422",
    pages = "7421--7432",
}

@inproceedings{
li2025:eagle3,
title={{EAGLE}-3: Scaling up Inference Acceleration of Large Language Models via Training-Time Test},
author={Yuhui Li and Fangyun Wei and Chao Zhang and Hongyang Zhang},
booktitle={The Thirty-ninth Annual Conference on Neural Information Processing Systems},
year={2025},
url={https://openreview.net/forum?id=4exx1hUffq}
}

@misc{Search-R1,
      title={Search-R1: Training LLMs to Reason and Leverage Search Engines with Reinforcement Learning}, 
      author={Bowen Jin and Hansi Zeng and Zhenrui Yue and Jinsung Yoon and Sercan Arik and Dong Wang and Hamed Zamani and Jiawei Han},
      year={2025},
      eprint={2503.09516},
      archivePrefix={arXiv},
      primaryClass={cs.CL},
      url={https://arxiv.org/abs/2503.09516}, 
}

@misc{sonnet3_5,
  title={Claude 3.5 Sonnet},
  author={Anthropic},
  year={2024},
  url={https://www.anthropic.com/news/claude-3-5-sonnet}
}

@inproceedings{xia:2024survey,
    title = "Unlocking Efficiency in Large Language Model Inference: A Comprehensive Survey of Speculative Decoding",
    author = "Xia, Heming  and
      Yang, Zhe  and
      Dong, Qingxiu  and
      Wang, Peiyi  and
      Li, Yongqi  and
      Ge, Tao  and
      Liu, Tianyu  and
      Li, Wenjie  and
      Sui, Zhifang",
    editor = "Ku, Lun-Wei  and
      Martins, Andre  and
      Srikumar, Vivek",
    booktitle = "Findings of the Association for Computational Linguistics: ACL 2024",
    month = aug,
    year = "2024",
    address = "Bangkok, Thailand",
    publisher = "Association for Computational Linguistics",
    url = "https://aclanthology.org/2024.findings-acl.456/",
    doi = "10.18653/v1/2024.findings-acl.456",
    pages = "7655--7671",
}

@article{Qwen2.5,
  author       = {An Yang and
                  Baosong Yang and
                  Beichen Zhang and
                  Binyuan Hui and
                  Bo Zheng and
                  Bowen Yu and
                  Chengyuan Li and
                  Dayiheng Liu and
                  Fei Huang and
                  Haoran Wei and
                  Huan Lin and
                  Jian Yang and
                  Jianhong Tu and
                  Jianwei Zhang and
                  Jianxin Yang and
                  Jiaxi Yang and
                  Jingren Zhou and
                  Junyang Lin and
                  Kai Dang and
                  Keming Lu and
                  Keqin Bao and
                  Kexin Yang and
                  Le Yu and
                  Mei Li and
                  Mingfeng Xue and
                  Pei Zhang and
                  Qin Zhu and
                  Rui Men and
                  Runji Lin and
                  Tianhao Li and
                  Tingyu Xia and
                  Xingzhang Ren and
                  Xuancheng Ren and
                  Yang Fan and
                  Yang Su and
                  Yichang Zhang and
                  Yu Wan and
                  Yuqiong Liu and
                  Zeyu Cui and
                  Zhenru Zhang and
                  Zihan Qiu},
  title        = {Qwen2.5 Technical Report},
  journal      = {CoRR},
  volume       = {abs/2412.15115},
  year         = {2024},
  url          = {https://doi.org/10.48550/arXiv.2412.15115},
  doi          = {10.48550/ARXIV.2412.15115},
  eprinttype   = {arXiv},
  eprint       = {2412.15115},
  bibsource    = {dblp computer science bibliography, https://dblp.org}
}

@misc{llama3,
      title={The Llama 3 Herd of Models}, 
      author={Aaron Grattafiori and Abhimanyu Dubey and Abhinav Jauhri and Abhinav Pandey and Abhishek Kadian and Ahmad Al-Dahle and others},
      year={2024},
      eprint={2407.21783},
      archivePrefix={arXiv},
      primaryClass={cs.AI},
      url={https://arxiv.org/abs/2407.21783}, 
}

@InProceedings{medusa,
  title = 	 {Medusa: Simple {LLM} Inference Acceleration Framework with Multiple Decoding Heads},
  author =       {Cai, Tianle and Li, Yuhong and Geng, Zhengyang and Peng, Hongwu and Lee, Jason D. and Chen, Deming and Dao, Tri},
  booktitle = 	 {Proceedings of the 41st International Conference on Machine Learning},
  pages = 	 {5209--5235},
  year = 	 {2024},
  editor = 	 {Salakhutdinov, Ruslan and Kolter, Zico and Heller, Katherine and Weller, Adrian and Oliver, Nuria and Scarlett, Jonathan and Berkenkamp, Felix},
  volume = 	 {235},
  series = 	 {Proceedings of Machine Learning Research},
  month = 	 {21--27 Jul},
  publisher =    {PMLR},
  url = 	 {https://proceedings.mlr.press/v235/cai24b.html},
}

@inproceedings{luo:2025tokenrecycle,
    title = "Turning Trash into Treasure: Accelerating Inference of Large Language Models with Token Recycling",
    author = "Luo, Xianzhen  and
      Wang, Yixuan  and
      Zhu, Qingfu  and
      Zhang, Zhiming  and
      Zhang, Xuanyu  and
      Yang, Qing  and
      Xu, Dongliang",
    editor = "Che, Wanxiang  and
      Nabende, Joyce  and
      Shutova, Ekaterina  and
      Pilehvar, Mohammad Taher",
    booktitle = "Proceedings of the 63rd Annual Meeting of the Association for Computational Linguistics (Volume 1: Long Papers)",
    month = jul,
    year = "2025",
    address = "Vienna, Austria",
    publisher = "Association for Computational Linguistics",
    url = "https://aclanthology.org/2025.acl-long.338/",
    doi = "10.18653/v1/2025.acl-long.338",
    pages = "6816--6831",
    ISBN = "979-8-89176-251-0",
}

@inproceedings{xia:2023specdec,
    title = "Speculative Decoding: Exploiting Speculative Execution for Accelerating Seq2seq Generation",
    author = "Xia, Heming  and
      Ge, Tao  and
      Wang, Peiyi  and
      Chen, Si-Qing  and
      Wei, Furu  and
      Sui, Zhifang",
    editor = "Bouamor, Houda  and
      Pino, Juan  and
      Bali, Kalika",
    booktitle = "Findings of the Association for Computational Linguistics: EMNLP 2023",
    month = dec,
    year = "2023",
    address = "Singapore",
    publisher = "Association for Computational Linguistics",
    url = "https://aclanthology.org/2023.findings-emnlp.257/",
    doi = "10.18653/v1/2023.findings-emnlp.257",
    pages = "3909--3925",
}

@inproceedings{gorilla,
 author = {Patil, Shishir G. and Zhang, Tianjun and Wang, Xin and Gonzalez, Joseph E.},
 booktitle = {Advances in Neural Information Processing Systems},
 doi = {10.52202/079017-4020},
 editor = {A. Globerson and L. Mackey and D. Belgrave and A. Fan and U. Paquet and J. Tomczak and C. Zhang},
 pages = {126544--126565},
 publisher = {Curran Associates, Inc.},
 title = {Gorilla: Large Language Model Connected with Massive APIs},
 url = {https://proceedings.neurips.cc/paper_files/paper/2024/file/e4c61f578ff07830f5c37378dd3ecb0d-Paper-Conference.pdf},
 volume = {37},
 year = {2024}
}

@InProceedings{LLMCompiler,
  title = 	 {An {LLM} Compiler for Parallel Function Calling},
  author =       {Kim, Sehoon and Moon, Suhong and Tabrizi, Ryan and Lee, Nicholas and Mahoney, Michael W. and Keutzer, Kurt and Gholami, Amir},
  booktitle = 	 {Proceedings of the 41st International Conference on Machine Learning},
  pages = 	 {24370--24391},
  year = 	 {2024},
  editor = 	 {Salakhutdinov, Ruslan and Kolter, Zico and Heller, Katherine and Weller, Adrian and Oliver, Nuria and Scarlett, Jonathan and Berkenkamp, Felix},
  volume = 	 {235},
  series = 	 {Proceedings of Machine Learning Research},
  month = 	 {21--27 Jul},
  publisher =    {PMLR},
  url = 	 {https://proceedings.mlr.press/v235/kim24y.html},
}

@inproceedings{zhu2025:dividethenaggregate,
  author       = {Dongsheng Zhu and
                  Weixian Shi and
                  Zhengliang Shi and
                  Zhaochun Ren and
                  Shuaiqiang Wang and
                  Lingyong Yan and
                  Dawei Yin},
  editor       = {Wanxiang Che and
                  Joyce Nabende and
                  Ekaterina Shutova and
                  Mohammad Taher Pilehvar},
  title        = {Divide-Then-Aggregate: An Efficient Tool Learning Method via Parallel
                  Tool Invocation},
  booktitle    = {Proceedings of the 63rd Annual Meeting of the Association for Computational
                  Linguistics (Volume 1: Long Papers), {ACL} 2025, Vienna, Austria,
                  July 27 - August 1, 2025},
  pages        = {28859--28875},
  publisher    = {Association for Computational Linguistics},
  year         = {2025},
  url          = {https://aclanthology.org/2025.acl-long.1401/},
  bibsource    = {dblp computer science bibliography, https://dblp.org}
}

@misc{xu2024conveyor,
      title={Conveyor: Efficient Tool-aware LLM Serving with Tool Partial Execution}, 
      author={Yechen Xu and Xinhao Kong and Tingjun Chen and Danyang Zhuo},
      year={2024},
      eprint={2406.00059},
      archivePrefix={arXiv},
      primaryClass={cs.CL},
      url={https://arxiv.org/abs/2406.00059}, 
}

@inproceedings{apibank,
    title = "{API}-Bank: A Comprehensive Benchmark for Tool-Augmented {LLM}s",
    author = "Li, Minghao  and
      Zhao, Yingxiu  and
      Yu, Bowen  and
      Song, Feifan  and
      Li, Hangyu  and
      Yu, Haiyang  and
      Li, Zhoujun  and
      Huang, Fei  and
      Li, Yongbin",
    editor = "Bouamor, Houda  and
      Pino, Juan  and
      Bali, Kalika",
    booktitle = "Proceedings of the 2023 Conference on Empirical Methods in Natural Language Processing",
    month = dec,
    year = "2023",
    address = "Singapore",
    publisher = "Association for Computational Linguistics",
    url = "https://aclanthology.org/2023.emnlp-main.187/",
    doi = "10.18653/v1/2023.emnlp-main.187",
    pages = "3102--3116",
}

@misc{nichols2025spec,
      title={Optimizing Agentic Language Model Inference via Speculative Tool Calls}, 
      author={Daniel Nichols and Prajwal Singhania and Charles Jekel and Abhinav Bhatele and Harshitha Menon},
      year={2025},
      eprint={2512.15834},
      archivePrefix={arXiv},
      primaryClass={cs.PL},
      url={https://arxiv.org/abs/2512.15834}, 
}


\appendix
\section{Details of \method}
\label{appendix:method_details}
In this section, we provide details of our proposed \method, including the implementation of parallel verification and the detailed algorithm.

\subsection{Parallel Verification}
Following prior work~\citep{eagle, medusa}, \method adopts a tree-based attention mechanism to verify multiple candidate draft sequences in parallel. Specifically, the candidate drafts are compacted into a single flattened sequence and fed into the target model $\mathcal{M}$. The model then performs a \textit{single forward pass} to verify all draft sequences simultaneously using a tree-structured attention mask, as illustrated in Figure~\ref{fig:intro}. 

To support different drafting strategies, \method employs two types of draft-tree architectures: (1) \textbf{Static trees.} For open-ended states (e.g., $q_v$ and $q_o$), \method adopts a general SD strategy (e.g., TR) for text generation, where a pre-defined static tree architecture is used for drafting; (2) \textbf{Dynamic trees.} For structural states (e.g., $q_t$ and $q_p$) and retrieval-augmented speculation, \method constructs dynamic tree masks that adapt to the length of each draft sequence, thereby reducing unnecessary verification computation.

\subsection{Algorithm}
We present the pseudo code of \method in Algorithm~\ref{alg:toolspec}. Concretely, at each decoding step, \method checks whether the current suffix matches any sequence in historical calls. If a match is found, it extracts the continuations as high-quality drafts. Otherwise, \method employs a finite-state machine (FSM) to guide schema-aware drafting, thereby accelerating tool-calling generation.
\section{Experimental Details}
\label{appendix:exp_details}
In this section, we present our experimental details, including dataset descriptions, implementation specifics, extended results, and additional analyses that complement our main findings.

\begin{algorithm}[tb]
\caption{Tool-calling with \method}
\label{alg:toolspec}
\small
\begin{algorithmic}[1]
\Require target model $\mathcal{M}$; input query $\boldsymbol{x}$; tool documentation $\mathcal{D}$; historical datastore $\mathcal{H}=\{(\boldsymbol{h}_j,\boldsymbol{y}_j)\}_{j=1}^{|\mathcal{H}|}$; parameter $k$.
\Ensure Generated tool-call sequence $\boldsymbol{y}$.

\State Parse documentation $\mathcal{D}$ into schema $\mathcal{S}$ %
\State Construct the FSM $\mathcal{F}=(\mathcal{Q}, \Sigma_{\text{schema}}, \delta, q_0)$ from $\mathcal{S}$
\State $\boldsymbol{y} \gets \emptyset$, $q \gets q_0$
\State Compute query representation $\boldsymbol{h}_x$
\State Retrieve historical calls $\mathcal{I}_k \gets \operatorname{Top-K}_{j} \text{sim}(\boldsymbol{h}_x,\boldsymbol{h}_j)$
\While{\textsc{NotFinished}$(q,\boldsymbol{y})$}
    \mydarkcolor{\LineComment{Retrieval-augmented speculation}}
    \State $\mathcal{B}_{\text{retrieval}} \gets \textsc{SuffixMatch}(\boldsymbol{y}, \{{\boldsymbol{y}}_{j}\}_{j \in \mathcal{I}_k})$
    \If{$\mathcal{B}_{\text{retrieval}} \neq \emptyset$}
        \State $\mathcal{B} \gets \mathcal{B}_{\text{retrieval}}$
    \Else
        \mydarkcolor{\DoubleLineComment{Schema-aware drafting}}
        \State $\mathcal{B} \gets 
        \textsc{SchemaAwareDraft}(q,\mathcal{S},\boldsymbol{y})$
    \EndIf
    \mydarkcolor{\LineComment{Packed sequence with tree attention}}
    \State $(\boldsymbol{z},\mathbf{A}) \gets \textsc{PackDraftsAsTree}(\mathcal{B})$
    \State $\hat{\mathcal{B}} \gets \textsc{Verification}(\mathcal{M}, \boldsymbol{x}, \boldsymbol{y}, \boldsymbol{z}, \mathbf{A})$
    \State $(\tilde{\boldsymbol{y}}^\star,\ell) \gets \textsc{SelectLongestAcceptedPrefix}(\hat{\mathcal{B}})$
    \State $\boldsymbol{y} \gets \boldsymbol{y} \oplus \tilde{\boldsymbol{y}}^\star_{1:\ell}$
    \mydarkcolor{\LineComment{Update FSM state}}
    \State $q \gets \delta(q, \tilde{\boldsymbol{y}}^\star_{1:\ell})$
\EndWhile

\State \Return $\boldsymbol{y}$
\end{algorithmic}
\end{algorithm}

\subsection{Datasets} 
\label{appendix:dataset_details}
Our evaluation encompasses four widely used tool-use benchmarks. The details of the evaluation data are illustrated as follows:
\begin{itemize}
    \item  \textbf{API-Bank}~\citep{apibank}: a three-level evaluation framework comprising 73 diverse APIs. It evaluates an LLM’s ability to select and invoke tools through natural multi-turn dialogues across varying levels of difficulty.
    \item  \textbf{ToolAlpaca}~\cite{ToolAlpaca}: a benchmark for assessing general-purpose tool-use capabilities of LLMs. The evaluation dataset consists of two subsets: (i) a simulated subset with 10 synthetic tools, and (ii) a real-world subset with 11 APIs from diverse domains.
    \item  \textbf{BFCLv2}~\cite{BFCL}: a comprehensive benchmark covering a wide range of scenarios, including single-step reasoning, multi-step tool use, real-time execution, etc. We evaluate \method under both single-function and multi-function settings.
    \item  \textbf{ToolBench}~\cite{Qin:2024toolllm}: a large-scale, high-quality benchmark featuring multi-tool and multi-turn interactions over 16,464 real-world APIs, automatically constructed using ChatGPT. We evaluate \method on its single-tool subset.
\end{itemize}

\subsection{Implementation Details} 
\label{appendix:impl_details}
The retrieval parameter $k$ is set to $10$, such that the top-$10$ candidates are retained at each retrieval step. For suffix matching, we follow PLD~\citep{saxena2023pld} and consider suffix lengths $L \in \{5,6,7\}$, applying a longest-first strategy (from $7$ to $5$). Once a match is found, we extract continuations following the matched suffix in sequential order, using predefined lengths $n \in \{32, 16, 8, 8\}$ to construct draft candidates. We adopt Token Recycling (TR)~\citep{luo:2025tokenrecycle} as the general speculative generation strategy in \method. All experiments were conducted using PyTorch 2.5.1 on 2$\times$NVIDIA A100-PCIE GPUs (40GB) with 40 CPU cores under CUDA 12.4. Inference for our method and all baselines was performed using the Hugging Face transformers package. Following prior work, we adopt speculative sampling~\citep{leviathan:2023spec} as our acceptance strategy, with a batch size of 1.

\subsection{Performance Comparison} 
\label{appendix:performance}
Table~\ref{tab:performance} illustrates the accuracy comparison between \method and vanilla autoregressive decoding across different models on API-Bank and ToolAlpaca. The results demonstrate that \method preserves the original performance of autoregressive decoding, achieving consistently comparable accuracy across various models and datasets. The slight discrepancy observed for Qwen2.5-14B on ToolAlpaca can be attributed to fp16 precision.

\begin{table}[t]
\centering
\small
\resizebox{\linewidth}{!}{
\begin{tabular}{@{}llccc@{}}
\toprule
\bf Models &\bf Methods & \bf API-Bank & \bf ToolAlpaca \\\midrule
\multirow{2}{*}{Qwen2.5-7B}
&Vanilla & 54.94 & 58.94 \\
&\method & \green{54.94} & \green{58.94} \\\midrule
\multirow{2}{*}{Qwen2.5-14B}
&Vanilla & 54.61 & 58.94 \\
&\method & \green{54.61} & \green{59.60} \\\midrule
\multirow{2}{*}{LLaMA-3.1-8B}
&Vanilla & 59.30 & 48.34 \\
&\method & \green{59.30} & \green{48.34} \\\midrule
\multirow{2}{*}{LLaMA-3.2-3B}
&Vanilla & 50.08 & 41.72 \\
&\method & \green{50.08} & \green{41.72} \\
\bottomrule
\end{tabular}}
\caption{Performance comparison between \method and vanilla autoregressive decoding across different models on two widely used tool-use benchmarks.}
\label{tab:performance}
\end{table}

\subsection{Hyperparameters in \method}
Table~\ref{tab:continuation} and Table~\ref{tab:topk} show the speedup of \method under different hyperparameter settings, including continuation length and the retrieval parameter $k$. The evaluation is conducted on 100 instances sampled from the training set of API-Bank~\citep{apibank}. Table~\ref{tab:continuation} shows that shorter continuation lengths (e.g., $\{16, 8, 4, 4\}$) lead to suboptimal \#MAT and speedup. Although longer continuation lengths (e.g., $\{64, 32, 16, 8\}$) slightly improve \#MAT, they also introduce additional computational overhead and reduce the overall speedup. Therefore, we adopt continuation lengths of $\{32, 16, 8, 8\}$ in our experiments. We further compare this setting with uniform continuation lengths of $\{16, 16, 16, 16\}$, and the results show that allocating longer continuations to top-ranked candidates brings greater benefits. Similarly, Table~\ref{tab:topk} shows that setting the retrieval parameter $k$ to 10 achieves the best speedup in our experiments.

\begin{table}[t]
\centering
\small
\begin{tabular}{@{}l|ccc@{}}
\toprule
\bf Continuations &\#MAT &Tokens/s & Speedup \\\midrule
$\{16, 8, 4, 4\}$ & 5.52 & 115.07 & 4.08$\times$  \\
$\{32, 16, 8, 8\}$ & 5.75 & \bf{124.77} & \bf{4.42$\times$} \\
$\{64, 32, 16, 8\}$ & \bf{5.77} & 122.60 & 4.35$\times$ \\
$\{16, 16, 16, 16\}$ & 5.60 & 117.95 & 4.18$\times$  \\
\bottomrule
\end{tabular}
\caption{Speedup of \method under different continuation lengths. Results are obtained with LLaMA-3.1-8B-Instruct on API-Bank.}
\label{tab:continuation}
\end{table}

\begin{table}[t]
\centering
\small
\begin{tabular}{@{}l|ccc@{}}
\toprule
\bf $k$ &\#MAT &Tokens/s & Speedup \\\midrule
$1$ & 5.13 & 113.18 & 4.01$\times$  \\
$5$ & 5.56 & 119.70 & 4.24$\times$ \\
$10$ & \bf{5.75} & \bf{124.77} & \bf{4.42$\times$} \\
$15$ & 5.69 & 123.22 & 4.37$\times$  \\
\bottomrule
\end{tabular}
\caption{Speedup of \method with different retrieval parameter values $k$. Results are obtained with LLaMA-3.1-8B-Instruct on API-Bank.}
\label{tab:topk}
\end{table}

\subsection{Memory Cost}
The additional memory overhead introduced by \method is modest. The primary persistent cost comes from the \textit{output memory}, which stores the hidden state of each input question,  along with its corresponding output tokens for each memorized sample. On API-Bank, this memory requires only 4.9 MB for 597 entries, corresponding to approximately 0.0082 MB per entry. To support larger datasets, \method can maintain a fixed-size memory bank, such as 1,000 memorized samples, thereby bounding the output memory while preserving the benefits of retrieval-augmented speculation. We leave the design of such memory management strategies for future investigation.

\begin{table}[t]
\centering
\small
\begin{tabular}{@{}l|ccc@{}}
\toprule
\textbf{Models} &\#MAT &Tokens/s & Speedup \\\midrule
Vanilla & 1.00 & 17.23 & 1.00$\times$  \\
PLD & 1.91 & 28.24 & 1.64$\times$ \\
TR & 2.90 & 30.30 & 1.76$\times$ \\
SAMD & \underline{2.99} & \underline{30.73} & \underline{1.78$\times$}  \\
\method & \bf{4.72} & \bf{52.42} & \bf{3.04$\times$}  \\
\bottomrule
\end{tabular}
\caption{Speedup of \method and prior plug-and-play methods using Qwen2.5-32B-Instruct on API-Bank. The best results are shown in \textbf{boldface}, and the second-best results are \underline{underlined}.}
\label{tab:32b}
\end{table}

\begin{figure}[t]
\centering
\includegraphics[width=0.95\columnwidth]{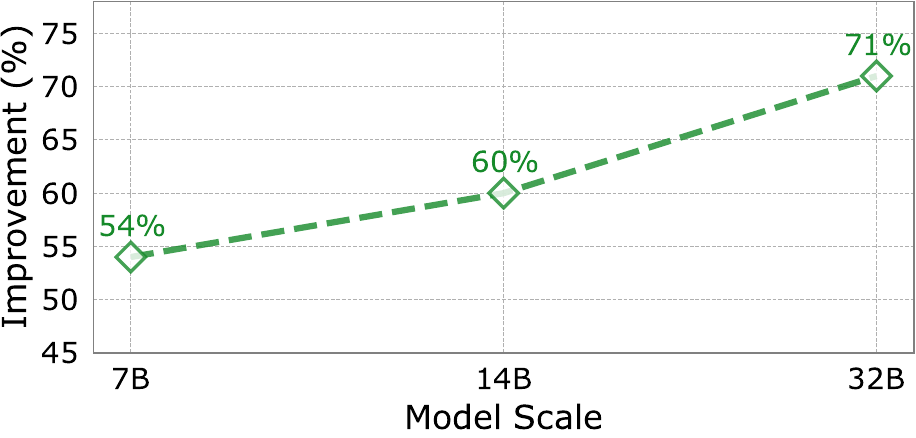}
\caption{Relative speedup improvement of \method over the previous best method. Results are obtained using the Qwen series on API-Bank.}
\label{fig:scaling}
\end{figure}

\subsection{Results on Qwen2.5-Instruct-32B}
\label{appendix:qwen32B}
Table~\ref{tab:32b} compares the speedup of \method with prior plug-and-play methods on API-Bank using Qwen2.5-32B-Instruct. The results demonstrate that \method outperforms these baselines and achieves up to a $3\times$ speedup over standard autoregressive decoding. Figure~\ref{fig:scaling} shows that the relative speedup gain of \method over the strongest reproduced baseline becomes larger as the model size increases. These results suggest that \method scales favorably with model size and provides increasingly pronounced efficiency benefits over existing plug-and-play approaches.

\begin{figure}[t]
\centering
\includegraphics[width=0.95\columnwidth]{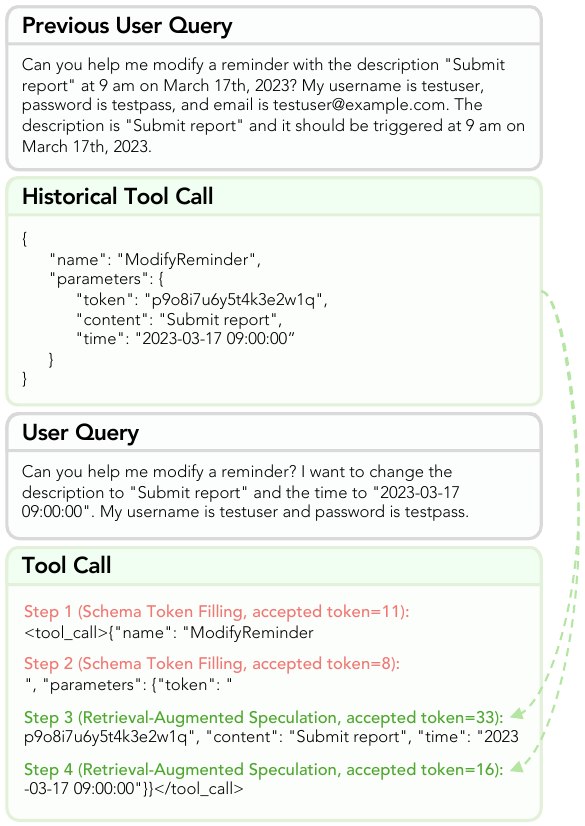}
\caption{Case study of \method with similar user queries on API-Bank using LLaMA-3.1-8B-Instruct.}
\label{fig:case1}
\end{figure}

\begin{figure}[t]
\centering
\includegraphics[width=0.95\columnwidth]{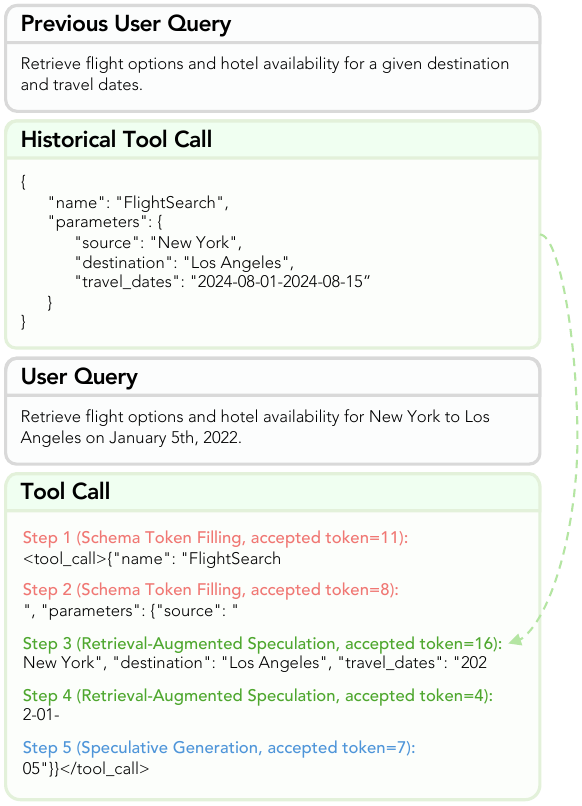}
\caption{Case study of \method under different tool-argument values using LLaMA-3.1-8B-Instruct.}
\label{fig:case2}
\end{figure}

\subsection{Case Study}
Figure~\ref{fig:case1} presents a case study of \method on API-Bank using LLaMA-3.1-8B-Instruct, where the current user query is similar to previously observed queries. As shown, \method first applies schema token filling to generate the required tool name and schema-constrained tokens. It then retrieves a highly similar historical tool call and reuses its continuation as draft tokens, yielding up to 33 accepted tokens in subsequent decoding steps. This case demonstrates the effectiveness of schema token filling and retrieval-augmented speculation.

Similarly, Figure~\ref{fig:case2} presents a case with similar user queries but different tool-argument values. \method reuses the continuation of a previous output as draft tokens, yielding 16 accepted tokens in the third decoding step. It then corrects the user-specific travel information and completes the tool call through general speculative generation. This case highlights the strength of \method in repeated tool-calling scenarios, where most tokens can be drafted from historical invocations while the target LLM focuses on key argument values.

\end{document}